%% file: iclr2023_conference_v2_arxiv.tex
\documentclass{article} 
\usepackage{iclr2023_conference,times}

\usepackage[hidelinks]{hyperref}
\usepackage{url}
\usepackage{booktabs}       
\usepackage{amsfonts}       
\usepackage{nicefrac}       
\usepackage{microtype}      
\usepackage{xcolor}         
\usepackage{graphicx}
\usepackage{amsmath}
\usepackage{amsfonts}
\usepackage{amssymb}
\usepackage{amsthm}
\usepackage{appendix}
\usepackage{algorithmicx}
\usepackage{algorithm}
\usepackage{algpseudocode} 
\usepackage{subfigure}
\usepackage{caption}
\usepackage{paralist}
\usepackage{tikz-network}
\usepackage{booktabs} 
\usepackage{subfigure}
\usepackage{multicol}
\usepackage[capitalise]{cleveref}
\usepackage{array}
\usepackage{wrapfig}
\usepackage{bm}

\newcolumntype{H}{>{\setbox0=\hbox\bgroup}c<{\egroup}@{}}

\newtheorem{theorem}{Theorem}

\def \E {\mathbb{E}}

\def \L {\mathcal{L}}

\def \R {\mathbb{R}}

\def \N {\mathcal{N}}

\def \Gh {\widehat{G}}

\def \Lh {\widehat{\L}}

\def \G {\mathcal{G}}

\def \Qh {\widehat{Q}}
\def \Gh {\widehat{\mathcal G}}
\def \G{\mathcal G}
\def \Ph {\widehat{P}}

\renewcommand{\vec}[1]{\bm {#1}}

\newcommand{\one}[0]{\bm {1}}

\renewcommand{\L}[0]{\mathcal L}

\newcommand{\til}[0]{\widetilde}

\newcommand{\onehot}[1]{\text{one-hot} \left ( #1 \right )}
\newcommand{\mlp}[2]{\text{MLP}_{\text{#1}}\!\left ( #2 \right )}
\newcommand{\pmt}[0]{\ensuremath{\pm}}

\Crefname{appendix}{App.}{Apps.}
\Crefname{equation}{Eq.}{Eqs.}
\Crefname{figure}{Fig.}{Figs.}
\Crefname{tabular}{Tab.}{Tabs.}
\Crefname{table}{Tab.}{Tabs.}
\Crefname{algorithm}{Alg.}{Algs.}
\Crefname{theorem}{Thm.}{Thms.}

\title{GLINKX: A Scalable Unified Framework For Homophilous and Heterophilous Graphs}

\author{Marios Papachristou\thanks{Work done while interning at Twitter.}\\
    Cornell University \\
    \texttt{papachristoumarios@gmail.com}
    \And
    Rishab Goel \\
    Twitter Inc. \\
    \texttt{rgoel@twitter.com}
    \And
    Frank Portman \\
    Twitter Inc. \\
    \texttt{fportman@twitter.com}
    \And
    Matthew Miller \\
    Twitter Inc. \\
    \texttt{mmiller@twitter.com}
    \And
    Rong Jin \\
    Twitter Inc. \\
    \texttt{rj@twitter.com}
}

\iclrfinalcopy 
\begin{document}

\maketitle

\begin{abstract}
    In graph learning, there have been two predominant inductive biases regarding graph-inspired architectures: On the one hand, higher-order interactions and message passing work well on homophilous graphs and are leveraged by GCNs and GATs. Such architectures, however, cannot easily scale to large real-world graphs. On the other hand,  shallow (or node-level) models using ego features and adjacency embeddings work well in heterophilous graphs. 
    In this work, we propose a novel scalable shallow method -- GLINKX -- that can work both on homophilous and heterophilous graphs. GLINKX leverages (i) novel monophilous label propagations (ii) ego/node features, (iii) knowledge graph embeddings as positional embeddings, (iv) node-level training, and (v) low-dimensional message passing.
    Formally, we prove novel error bounds and justify the components of GLINKX. Experimentally, we show its effectiveness on several homophilous and heterophilous datasets.
\end{abstract}

\section{Introduction} \label{sec:introduction}
In recent years, graph learning methods have emerged with a strong performance for various ML tasks. Graph ML methods leverage the topology of graphs underlying the data \citep{battaglia2018relational} to improve their performance. Two very important design options for proposing graph ML based architectures in the context of \emph{node classification} are related to whether the data is \emph{homophilous} or \emph{heterophilous}. 

For homophilous data -- where neighboring nodes share similar labels \citep{mcpherson2001birds, altenburger2018monophily} -- Graph Neural Network (GNN)-based methods are able to achieve high accuracy. Specifically, a broad subclass sucessfull GNNs are Graph Convolutional Networks (GCNs)  (e.g., GCN, GAT, etc.) \citep{kipf2016semi,velivckovic2017graph,zhu2020beyond}. In the GCN paradigm, \emph{message passing} and \emph{higher-order interactions} help node classification tasks in the homophilous setting since such inductive biases tend to bring the learned representations of linked nodes close to each other. However, GCN-based architectures suffer from \emph{scalability issues}. Performing (higher-order) propagations during the training stage are hard to scale in large graphs because the number of nodes grows exponentially with the increase of the filter receptive field. Thus, for practical purposes, GCN-based methods require \emph{node sampling}, substantially increasing their training time. For this reason, architectures \citep{huang2020combining, zhang2022graph, sun2021scalable, maurya2021improving, rossi2020sign} that leverage propagations outside of the training loop (as a preprocessing step) have shown promising results in terms of scaling to large graphs. 

In \emph{heterophilous} datasets \citep{rogers2014diffusion}, the nodes that are connected tend to have different labels. Currently, many works that address heterophily can be classified into two categories concerning scale. On the one hand, recent successful architectures (in terms of accuracy) \citep{jin2022raw, di2022graph, zheng2022graph, luan2021heterophily, chien2020adaptive, lei2022evennet} that address heterophily resemble GCNs in terms of design and thus suffer from the same scalability issues. On the other hand, \emph{shallow or node-level models} (see, e.g., \citep{lim2021large, zhong2022simplifying}), i.e., models that are treating graph data as tabular data and do not involve propagations during training, has shown a lot of promise for large heterophilous graphs. In \citep{lim2021large}, it is shown that combining ego embeddings (node features) and adjacency embeddings works in the heterophilous setting. One element that LINKX exploits via the adjacency embeddings is monophily \citep{altenburger2018monophily, altenburger2018node}, namely the similarity of the labels of a node's neighbors.
However, their design is still impractical in real-world data since the method (LINKX) is \emph{not} inductive (see \cref{sec:preliminaries}), and embedding the adjacency matrix directly requires many parameters in a model. 
In LINKX, the adjacency embedding of a node can alternatively be thought of as a \emph{positional embedding} (PE) of the node in the graph, and recent developments \citep{kim2022pure, dwivedi2021graph, lim2021large} have shown the importance of PEs in both homophilous and heterophilous settings. 
However, most of these works suggest PE parametrizations that are difficult to compute in large-scale settings. We argue that PEs can be obtained in a scalable manner by utilizing \emph{knowledge graph embeddings}, which, according to prior work, can  \citep{el2022twhin, lerer2019pytorch, bordes2013translating, yang2014embedding} be trained in very large networks.

\noindent \textbf{Goal \& Contribution:} In this work, we develop a scalable method for node classification that: (i) works both on homophilous and heterophilous graphs, (ii) is \emph{simpler and faster} than conventional message passing networks (by avoiding the neighbor sampling and message passing overhead during training), and (iii) can work in both a \emph{transductive} and an \emph{inductive}\footnote{For this paper, we operate in the \emph{transductive setting}. See \cref{app:inductive} for the inductive setting.} setting. For a method to be scalable, we argue that it should: (i) run models on node-scale (thus leveraging i.i.d. minibatching), (ii) avoid doing message passing during training and do it a constant number of times before training, and (iii) transmit small messages along the edges. Our proposed method -- GLINKX (see \cref{sec:glinkx}) -- combines all the above desiderata. GLINKX has three components: (i) ego embeddings \footnote{We use ego embeddings and node features interchangeably.}, (ii) PEs inspired by architectures suited for heterophilous settings, and (iii) \emph{scalable} 2nd-hop-neighborhood  propagations inspired by architectures suited for monophilous settings~(\cref{sec:phillies}). We provide novel theoretical error bounds and justify components of our method (\cref{sec:theoretical_analysis}). Finally, we evaluate GLINKX's empirical effectiveness on several homophilous and heterophilous datasets~(\cref{sec:experiments}).

\section{Preliminaries} \label{sec:preliminaries}

\subsection{Notation}

We denote scalars with lower-case, vectors with bold lower-case letters, and matrices with bold upper-case. We consider a directed graph $G = G(V, E)$ with vertex set $V$ with $|V| = n$ nodes, and edge set $E$ with $|E| = m$ edges, and adjacency matrix $\vec{A}$. $\vec{X} \in \mathbb R^{n \times d_X}$ represents the $d_X$-dimensional features and $\vec{P} \in \mathbb R^{n \times d_P}$ represent the $d_P$-dimensional PE matrix. A node $i$ has a feature vector $\vec{x}_i \in \mathbb R^{d_X}$ and a positional embedding $\vec{p}_i \in \mathbb R^{d_P}$ and belongs to a class $y_i \in \{1, \dots, c \}$. {The training set is denoted by $G_{\text{train}}(V_{\text{train}}, E_{\text{train}})$, the validation set by $G_{\text{valid}}(V_{\text{valid}}, E_{\text{valid}})$, and test set by $G_{\text{test}}(V_{\text{test}}, E_{\text{test}})$.} $\mathbb I \{ \cdot \}$ is the indicator function. $\Gamma_c$ is the $c$-dimensional simplex. 

\begin{algorithm}[t]

\caption{GLINKX Algorithm} 
\label{alg:glinkx}
{\begin{flushleft}
\textbf{Input:} Graph $G(V, E)$ with train set $V_{\text{train}} \subseteq V$, node features $\vec X$, labels $\vec Y$ \\
\textbf{Output:} Node Label Predictions $\vec Y_{\text{final}}$ \\

\textbf{1st Stage (KGEs).} Pre-train knowledge graph embeddings $\vec P$ with Pytorch Biggraph. 

\textbf{2nd Stage (MLaP).} Propagate labels and predict neighbor distribution
\begin{compactenum}
    \item \textbf{MLaP Forward:} Calculate {the distribution of each training node's neighbors, i.e.} $\vec {\hat y}_i = \frac {\sum_{j \in V_{\text{train}} : (j, i) \in E_{\text{train}}} \vec y_j} {| \{ j \in V_{\text{train}}: (j, i) \in E_{\text{train}} \}|}$  for all $i \in V_{\text{train}}$
    \item \textbf{Learn distribution of a node's neighbors:} 
    \begin{compactenum}
        \item For each epoch, calculate  $\vec {\tilde y}_i = f_1(\vec x_i, \vec p_i; \vec \theta_1)$ for $i \in V_{\text{train}}$
        \item Update the parameters s.t. the {negative} cross-entropy $\mathcal L_{\text{CE}, 1} (\vec \theta_1) = \sum_{i \in V_{\text{train}}} \text{CE}(\vec {\hat y}_i, \vec {\tilde y}_i; \vec \theta_1)$ is maximized {in order to bring $\vec {\tilde y_i}$ statistically close to $\vec {\hat y_i}$}. 
        
        \item Let $\vec \theta_1^*$ be the parameters at the end of the training that correspond to the epoch with the best validation accuracy.
    \end{compactenum}
    
    \item \textbf{MLaP Backward:} Calculate $\vec {y}_i' = \frac {\sum_{j \in V : (i, j) \in E} \vec {\tilde y}_j} {| \{ j \in V : (i, j) \in E \}|}$ for all $i \in V_{\text{train}}$, where $\vec {\tilde y}_j = f_1(\vec x_j, \vec p_j; \vec \theta_1^*)$.
\end{compactenum}

\textbf{3rd Stage (Final Model).} {Predicting node's own label distribution:} 
\begin{compactenum}
    \item For each epoch, calculate $y_{\text{final}, i} = f_2(\vec x_i, \vec p_i, \vec {y}_i'; \vec \theta_2)$.
    \item Update the parameters s.t. the {negative} cross-entropy $\mathcal L_{\text{CE}, 2} (\vec \theta_2) = \sum_{i \in V_{\text{train}}} \text{CE}(y_i, y_{\text{final}, i}; \vec \theta_2)$ is maximized.
\end{compactenum}

Return $\vec Y_{\text{final}}$

\end{flushleft}
}
\end{algorithm}

\subsection{Graph Convolutional Neural Networks}

In homophilous datasets, GCN-based methods have been used for node classification. GCNs \citep{kipf2016semi} utilize feature propagations together with non-linearities to produce node embeddings. Specifically, a GCN consists of multiple layers where each layer $i$ collects $i$-th hop information from the nodes through propagations and forwards this information to the $i + 1$-th layer. More specifically, if $G$ has a symmetrically-normalized adjacency matrix $\vec A_{sym}'$ (with self-loops) (ignoring the directionality of edges), then a GCN layer has the form

\begin{equation*}
\begin{split}
    \vec H^{(0)}  = \vec X, \; \vec H^{(i + 1)}  = \sigma \left ( \vec A_{sym}' \vec H^{(i)} \vec W^{(i)} \right ) \forall i \in [L], \; \vec Y  = \mathrm{softmax} \left ( \vec H^{(L)} \right ).
\end{split}
\end{equation*}

Here $\vec H^{(i)}$ is the embedding from the previous layer, $\vec W^{(i)}$ is a learnable projection matrix and $\sigma (\cdot)$ is a non-linearity (e.g. ReLU, sigmoid, etc.). 

\subsection{LINKX}

In heterophilous datasets, the simple method of LINKX has been shown to perform well. LINKX combines two components -- MLP on the node features $\vec X$ and LINK regression \citep{altenburger2018monophily} on the adjacency matrix -- as follows:

\begin{equation*}
\begin{split}
    \vec H_X = \mlp {$X$} {\vec X}, \; \vec H_A  = \mlp {$A$} {\vec A}, \; \vec Y  = \mathrm{ResNet}(\vec H_X, \vec H_A).
\end{split} 
\end{equation*}

\subsection{Node Classification} 

In node classification problems on graphs, we have a model $f(\vec X, \vec Y_{\text{train}}, \vec A; \vec \theta)$ that takes as an input the node features $\vec X$, the training labels $\vec Y_{\text{train}}$ and the graph topology $\vec A$ and produces a prediction for each node $i$ of $G$, which corresponds to the probability that a given node belongs to any of $c$ classes (with the sum of such probabilities being one). The model is trained with back-propagation. Once trained, the model can be used for the prediction of labels of nodes in the test set.

There are two training regimes: \emph{transductive} and \emph{inductive}. In the transductive training regime, we have full knowledge of the graph topology (for the train, test, and validation sets) and the node features, and the task is to predict the labels of the validation and test set. In the inductive regime, only the graph induced by $V_{\text{train}}$ is known at the time of training, and then the full graph is revealed for prediction on the validation and test sets. In real-world scenarios, such as online social networks, the dynamic nature of problems makes the inductive regime particularly useful.

\subsection{Homophily, Heterophily \& Monophily}\label{sec:phillies}

\noindent \textbf{\emph{Homophily and Heterophily:}} There are various measures of homophily in the GNN literature like node homophily and edge homophily \cite{lim2021large}. Intuitively, homophily in a graph implies that nodes with similar labels are connected. GNN-based approaches like GCN, GAT, etc., leverage this property to improve the node classification performance. Alternatively, if a graph has low homophily -- namely, nodes that connect tend to have different labels -- it is said to be \emph{heterophilous}. In other words, a graph is heterophilous if neighboring nodes do not share similar labels. 

\noindent \textbf{\emph{Monophily:}} Generally, we define a graph to be monophilous if the label of a node is similar to that of its neighbors' neighbors\footnote{A similar definition of monophily has appeared in \citep{altenburger2018monophily},  whereby many nodes have extreme preferences for connecting to a certain
class.}. {Etymologically, the word ``monophily'' is derived from the Greek words \emph{``monos''} (unique) and \emph{``philos''} (friend), which in our context means that a node  -- regardless of its label -- has neighbors of primarily one label.}
In the context of a directed graph, monophily can be thought of as a structure that resembles \cref{subfig:glinkx_2} where similar nodes (in this case, three green nodes connected to a yellow node) are connected to a node with a different label. 

{We argue that encoding monophily into a model \textbf{can be helpful for both heterophilous and homophilous graphs (see \cref{subfig:homophilous,subfig:heterophilous}), which is one of the main motivators behind our work}.} In homophilous graphs, monophily will fundamentally encode the 2nd-hop neighbor's label information, and since in such graphs, neighboring nodes have similar labels, it can provide a helpful signal for node classification. In heterophily, neighboring nodes have different labels, but the  2nd-hop neighbors may share the same label, providing helpful information for node classification. Monophily is effective for heterophilous graphs \citep{lim2021large}. Therefore, an approach encoding monophily has an advantage over methods designed specifically for homophilous and heterophilous graphs, especially when varying levels of homophily can exist between different sub-regions in the same graph (see \cref{sec:varying_homophily}). It may also not be apparent if the (sub-)graph is purely homophilous/heterophilous (since these are not binary constructs), which makes a unified architecture that can leverage graph information for both settings all the more important.

\section{Method} \label{sec:glinkx}

\subsection{Components \& Motivation} \label{sec:components}

The desiderata we laid down on \cref{sec:introduction} can be realized by three components: (i) PEs, (ii) ego embeddings, and (iii) label propagations that encode monophily. More specifically, ego embeddings and PEs are used as primary features, which have been shown to work for both homophilous and heterophilous graphs for the models we end up training. Finally, the propagation step is used to encode monophily to provide additional information to our final prediction. 

\begin{figure*}
    \centering
    \subfigure[1st Stage\label{subfig:glinkx_1_block}]{\includegraphics[width=0.45\textwidth]{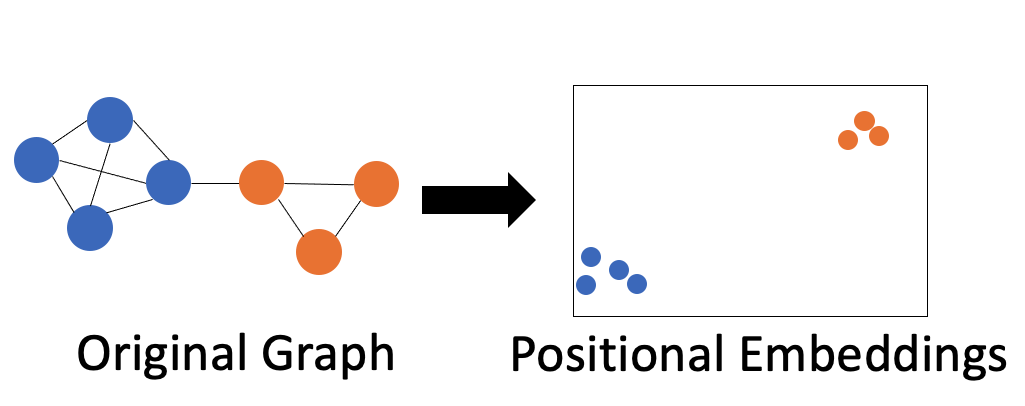}} \\
    \subfigure[2nd Stage\label{subfig:glinkx_2_block}]{ \includegraphics[width=0.45\textwidth]{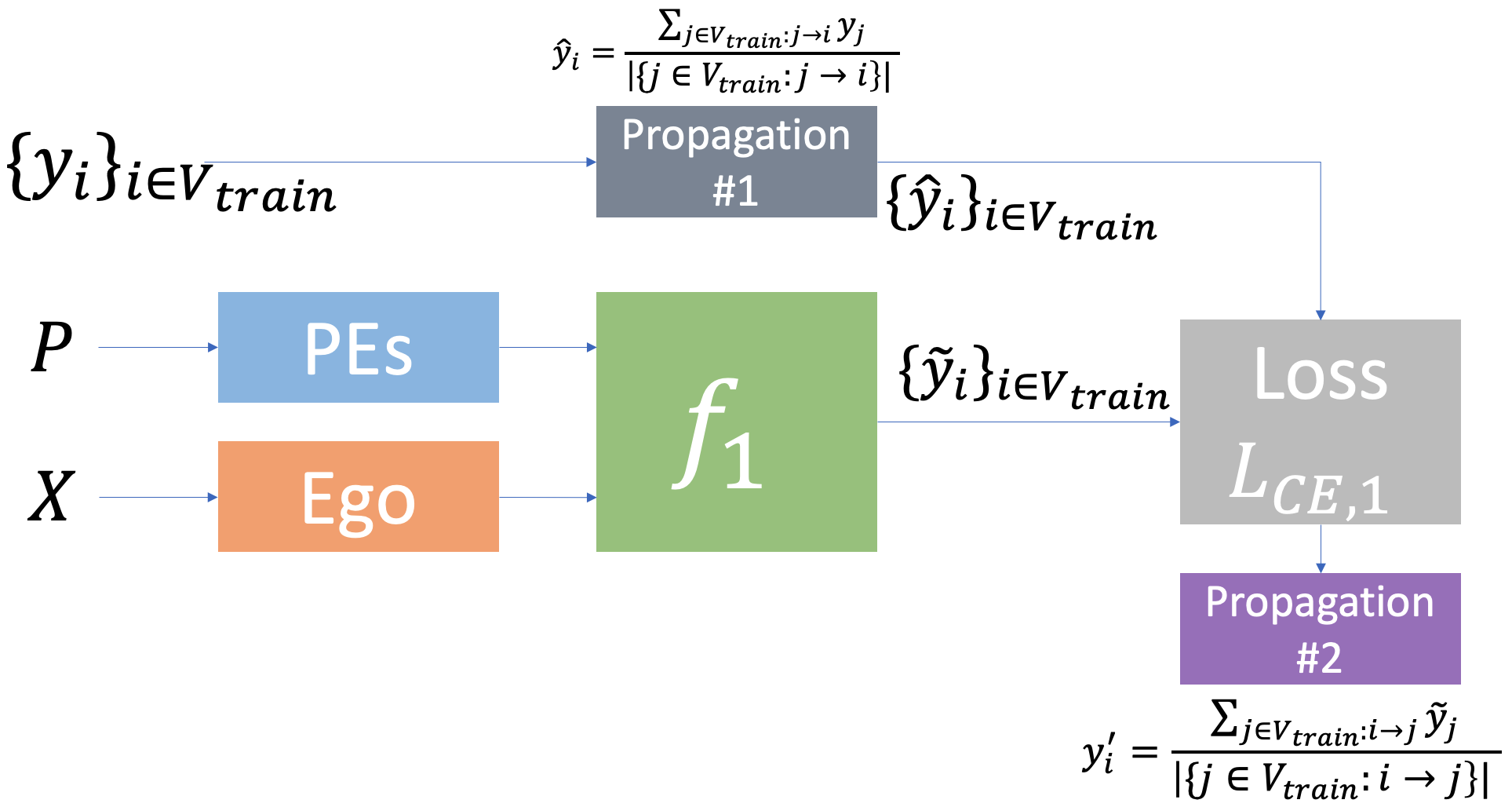}}
    \subfigure[3rd Stage\label{subfig:glinkx_3_block}]{ \includegraphics[width=0.45\textwidth]{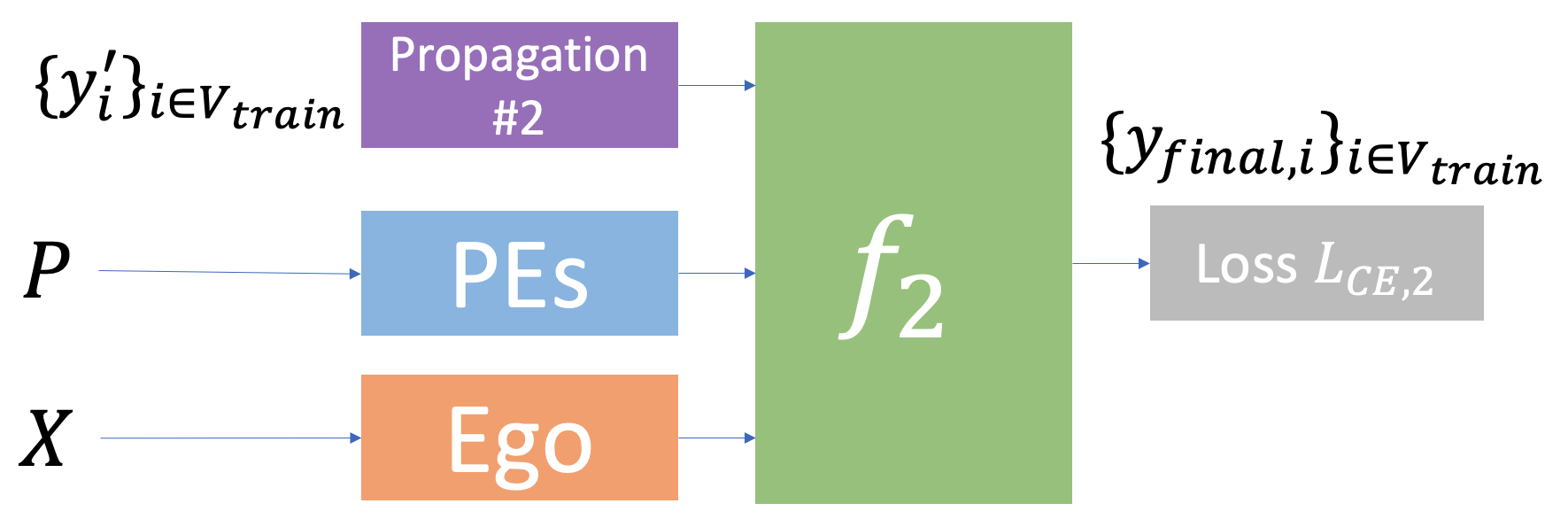}}
    \vspace{-\baselineskip}
    \caption{{Block Diagrams of GLINKX stages.}}
    \vspace{-\baselineskip}
    \label{fig:stages_block}
\end{figure*}

\textbf{\emph{Positional Embeddings:}}  We use PEs to provide our model information about the position of each node and hypothesize that PEs are an important piece of information in the context of large-scale node classification. PEs have been used to help discriminate isomorphic graph (sub)-structures \citep{kim2022pure, dwivedi2021graph, srinivasan2019equivalence}. This is useful for both homophily \citep{kim2022pure, dwivedi2021graph} and heterophily \citep{lim2021large} because isomorphic (sub)-structures can exist  in both the settings. In the homophilous case, adding positional information can help distinguish nodes that have the same neighborhood but distinct position \citep{dwivedi2021graph, morris2019weisfeiler, xu2018how}, circumventing the need to do higher-order propagations \citep{dwivedi2021graph, li2019deepgcns, bresson2017residual} which are prone to over-squashing \citep{alon2021on}. In heterophily, structural similarity among nodes is important for classification, as in the case of LINKX -- where adjacency embedding can be considered a PE. However, in large graphs, using adjacency embeddings or Laplacian eigenvectors (as methods such as \citep{kim2022pure} suggest) can be a computational bottleneck and may be infeasible. 
 
In this work, we leverage \emph{knowledge graph embeddings} (KGEs) to encode positional information about the nodes, {and embed the graph}. Using KGEs has two benefits: Firstly,  KGEs can be trained quickly for large graphs. This is because KGEs compress the adjacency matrix into a fixed-sized embedding, and adjacency matrices have been shown to be effective in heterophilous cases. Further, KGEs are lower-dimensional than the adjacency matrix (e.g., $d_P \sim 10^2$), allowing for faster training and inference times. Secondly, KGEs can be pre-trained efficiently on such graphs \citep{lerer2019pytorch} and can be used off-the-shelf for other downstream tasks, including node classification \citep{el2022twhin}\footnote{{Positional information can also be provided by other  methods such as node2vec \citep{grover2016node2vec}, however, most of such methods are less scalable.}}.  So, in the 1st Stage of our methods in \cref{alg:glinkx} (\cref{subfig:glinkx_1_block}) we train KGEs model on the available graph structure. Here, we fix this positional encoding once they are pre-trained for downstream usage. Finally, we note that this step is transductive but we can easily make it inductive \citep{el2022twhin, albooyeh2020out}. 

\textbf {\emph{Ego Embeddings:}} We get ego embeddings from the node features. Such embeddings have been used in homophilous and heterophilous settings \citep{lim2021large, zhu2020beyond}. Node embeddings are useful for tasks where the graph structure provides little/no information about the task. 

\textbf {\emph{Monophilous Label Propagations:}} We now propose a novel monophily (refer \cref{sec:phillies}) inspired label propagation which we refer to as Monophilous Label Propagation (MLaP). MLaP has the advantage that we can use it both for homophilous and heterophilous graphs or in a scenario with varying levels of graph homophily (see  \cref{sec:varying_homophily}) as it encodes monophily (\cref{sec:phillies}). 

To understand how MLaP encodes monophily, we consider the example in \cref{fig:glinkx}. In this example, we have three green nodes connected to a yellow node and two nodes of different colors connected to the yellow node. Then, one way to encode monophily in \cref{subfig:glinkx_2} while predicting label for $j_\ell, \ell \in [5]$,  is to get a \emph{distribution} of labels of nodes connected to node $i$ thus encoding its neighbors' distribution. The fact that there are more nodes with green color than other colors can be used by the model to make a prediction. But this information may only sometimes be present, or there may be few labeled nodes around node $i$. Consequently, we propose to use a model that predicts the label distribution of nodes connected to $i$. We use the node features ($\vec{x}_i$) and PE ($\vec{p}_i$) of node $i$ to build this model since nodes that are connected to node $i$ share similar labels, and thus, the features of node $i$ must be predictive of its neighbors. So, in \cref{subfig:glinkx_2}, we train a model to predict a distribution of $i$'s neighbors. Next, we provide $j_\ell$ the learned distribution of $i$'s neighbors by propagating the learned distribution from $i$ back to $j_\ell$. \cref{eq:mlap1,eq:mlap2,eq:mlap3} correspond to MLaP. We train a final model that leverages this information together with node features and PEs (\cref{subfig:glinkx_3}). 


\subsection{Our Method: GLINKX} \label{sec:method}

\begin{figure*}
    \centering
    \subfigure[\label{subfig:glinkx_2}MLaP Forward \& Neighbor Model]{\includegraphics[width=0.4\textwidth]{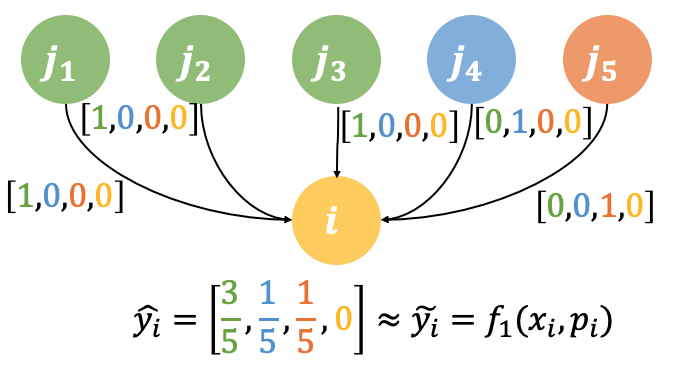}} 
    \subfigure[\label{subfig:glinkx_3}MLaP Backward \& Final Model]{\includegraphics[width=0.4\textwidth]{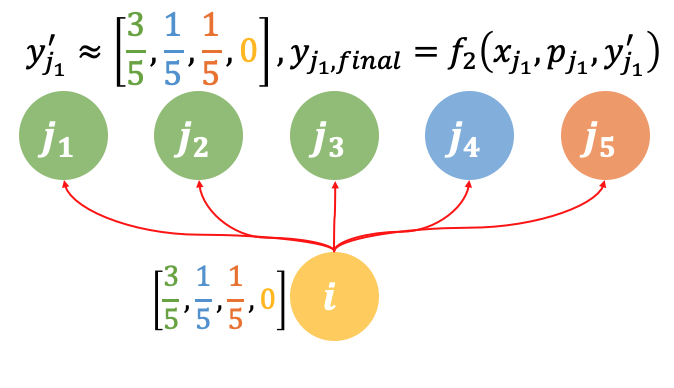}}
    \vspace{-\baselineskip}
    \caption{Example. For node $i$ we want to learn a model that takes $i$'s features $\vec{x}_i \in \mathbb R^{d_X}$, and PEs $\vec{p}_i \in \mathbb R^{d_P}$ and predict a value $\til {\vec{y}}_i \in \mathbb R^c$ that matches the label distribution of it's neighbors neighbors $\vec {\hat y}_i$ using a shallow model. Next, we want to propagate (outside the training loop) the (predicted) distribution of a node back to its neighbors and use it together with the ego features and the PEs to make a prediction about a node's own label. We propagate $\vec {\tilde y}_i$ to its neighbors $j_1$ to $j_5$. For example, for $j_1$, we encode the propagated distribution estimate $\vec {\tilde y}_i$ from $i$ to form $\vec y_{j_1}'$. We predict the label by using $\vec y_{j_1}', \vec{x}_{j_1}, \vec{p}_{j_1}$.}
    \label{fig:glinkx}
    \vspace{-1\baselineskip}
\end{figure*}

{We put the components discussed in \cref{sec:components} together into three stages. In the first stage, we pre-train the PEs by using KGEs. Next, encode monophily into our model by training a model that predicts a node's neighbors' distribution and by propagating the soft labels from the fitted model. Finally, we combine the propagated information, node features, and PEs to train a final model. GLINKX is described in \cref{alg:glinkx} and consists of three main components detailed as block diagrams in \cref{fig:stages_block}. \cref{fig:glinkx} shows the GLINKX stages from \cref{alg:glinkx} on a toy graph:}

\underline{\emph{1st Stage (KGEs):}} We train DistMult KGEs with Pytorch-Biggraph \citep{yang2014embedding} {treating $G$ as a knowledge graph with only one relation (see \cref{sec:kge_training} for more details)}.  {Here we have decided to use DistMult, but one can use their method of choice to embed the graph.}

{\underline{\emph{2nd Stage (MLaP):}}} First (2nd Stage in \cref{alg:glinkx}, \cref{subfig:glinkx_2_block}, and \cref{subfig:glinkx_2}), for a node we want to learn the distribution of \emph{its neighbors}. To achieve this, we propagate the labels from a node's neighbors (we call this step MLaP Forward), i.e. calculate 

\begin{equation} \label{eq:mlap1}
    \vec {\hat y}_i = \frac {\sum_{j \in V_{\text{train}} :(j, i) \in E_{\text{train}}} \vec y_j} {| \{ j \in V_{\text{train}}: (j, i) \in E_{\text{train}} \}|} \qquad \forall i \in V_{\text{train}}.
\end{equation}

Then, we train a model that predicts the distribution of neighbors, which we denote with $\vec {\tilde y}_i$ using the ego features $\{ \vec x_i \}_{i \in V_{\text{train}}}$ and the PEs $\{ \vec p_i \}_{i \in V_{\text{train}}}$ and maximize the {negative} cross-entropy with  treating $\{ \vec {\hat y}_i \}_{i \in V_{\text{train}}}$ as ground truth labels, namely we maximize

\begin{equation} \label{eq:mlap2}
    \mathcal L_{\text{CE, 1}}(\vec \theta_1) = \sum_{i \in V_{\text{train}}} \sum_{l \in [c]} \vec {\hat y}_{i,l} \log (\vec {\tilde y}_{i, l}),
\end{equation}

where  $\vec {\tilde y}_{i} = f_1 (\vec x_i, \vec p_i; \vec \theta_1)$ and $\vec \theta_1 \in \Theta_1$ is a learnable parameter vector. Although in this paper we assume to be in the \emph{transductive setting}, this step allows us to be inductive (see \cref{app:inductive}). In \cref{sec:theoretical_analysis} we give a theoretical justification of this step, namely \emph{``why is it good to use a parametric model to predict the distribution of neighbors?''}.

{Finally, we propagate the predicted soft-labels $\vec {\tilde y}_i$ back to the original nodes, i.e. calculate}

{
\begin{equation} \label{eq:mlap3}
    \vec y_i' = \frac {\sum_{j \in V : (i, j) \in E} \vec {\tilde y}_j} {| \{ j \in V : (i, j) \in E \}|} \quad \forall i \in V_{\text{train}},
\end{equation}
}

{where the soft labels $\{ \vec {\tilde y}_i \}_{i \in V_{\text{train}}}$ have been computed with the parameter $\vec \theta_1^*$ of the epoch with the best validation accuracy from model $f_1(\cdot | \vec \theta_1)$. We call this step MLaP Backward.}

{\underline{\emph{3rd Stage (Final Model):}}  We make the final predictions $\vec y_{\text{final, i}} = f_2(\vec x_i, \vec p_i, \vec y_i' ; \vec \theta_2)$ by combining the ego embeddings, PEs, and the (back)-propagated soft labels ($\vec \theta_2$ is a learnable parameter vector). We use the soft-labels $\vec {\tilde y}_i$ instead of the actual labels $\onehot {y_i}$ in order to avoid label leakage, which hurts performance (see also \citep{shi2020masked} for a different way to combat label leakage).}
{Finally, we maximize the {negative} cross-entropy with respect to a node's own labels,}

{
\begin{equation}
    \mathcal L_{\text{CE, 2}}(\vec \theta_2)  = \sum_{i \in V_{\text{train}}} \sum_{l \in [c]} \mathbb I \{ y_i = l \}  \log (\vec y_{\text{final, i}, l}),
\end{equation}
}

Overall, Stage 2 corresponds to learning the neighbor distributions and propagating these distributions, and Stage 3 uses these distributions to train a new model which predicts a node's labels. In \cref{sec:theoretical_analysis}, we prove that such a two-step procedure incurs lower errors than directly using the features to predict a node's labels.

\textbf{\emph{Scalability:}} GLINKX is highly scalable as it performs message passing a constant number of times by paying an $O(mc)$ cost, where the dimensionality of classes $c$ is usually small (compared to $d_X$ that GCNs rely on). In both Stages 2 and 3 of \cref{alg:glinkx}, we train node-level MLPs, which allow us to leverage i.i.d. (row-wise) mini-batching, like tabular data, and thus our complexity is similar to other shallow methods (LINKX, FSGNN) \citep{lim2021large, maurya2021improving}. This, combined with the propagation outside the training loops, circumvents the scalability issues of GCNs. For more details, refer \cref{sec:scalability}.

\begin{figure}
    \centering
    \subfigure[Node and class homophily\label{subfig:distribution}]{\includegraphics[width=0.5\textwidth]{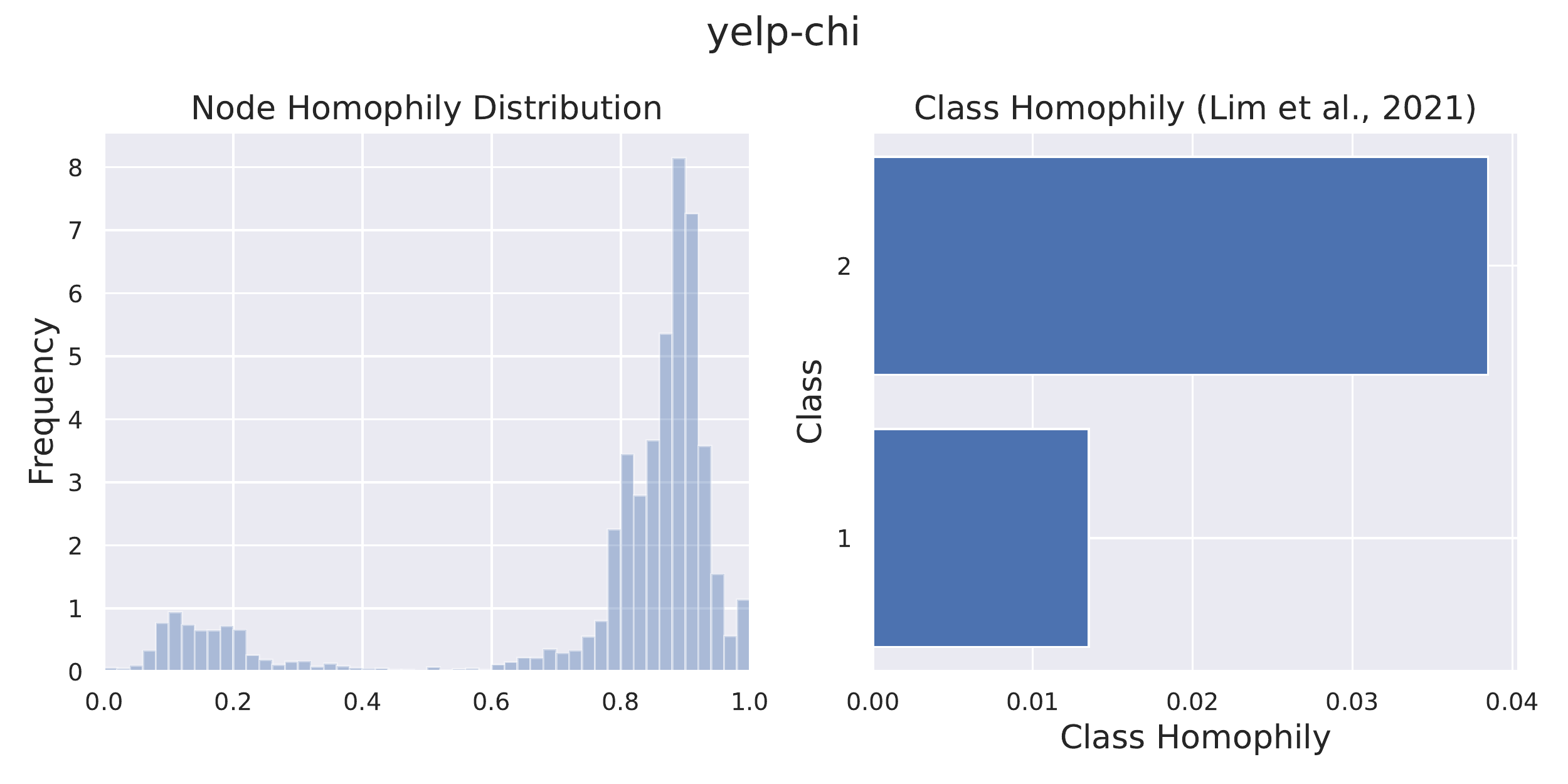}} \\ 
    \subfigure[Homophilous example\label{subfig:homophilous}]{\includegraphics[width=0.25\textwidth]{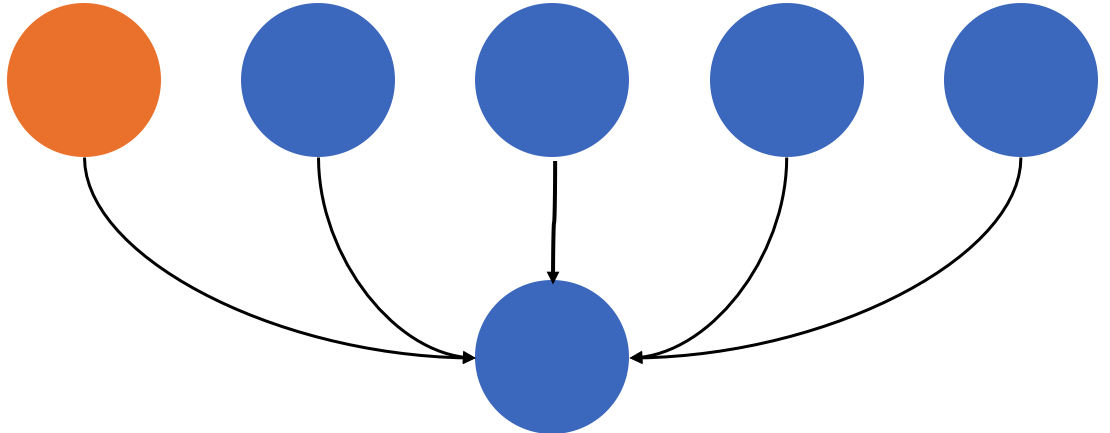}}
    \subfigure[Heterophilous example\label{subfig:heterophilous}]{\includegraphics[width=0.25\textwidth]{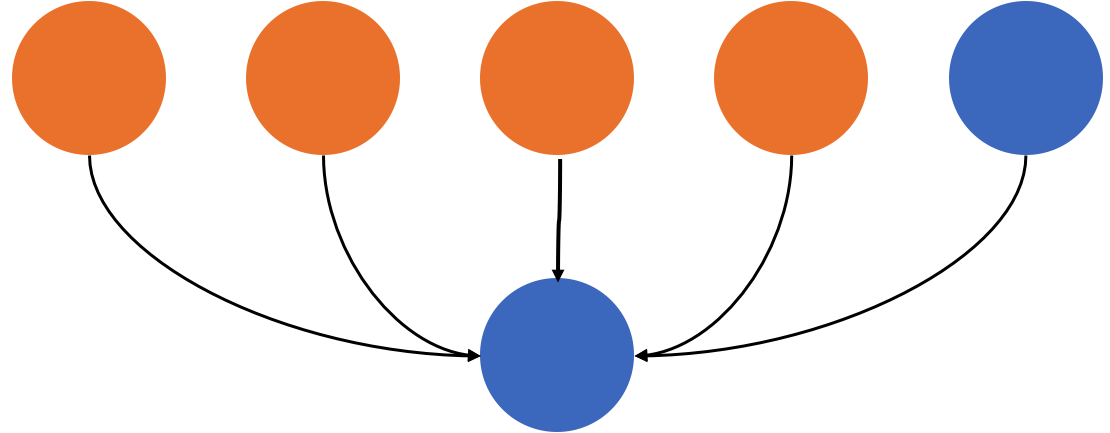}}
    \vspace{-1\baselineskip}
    \caption{Top: Node and class homophily distributions for the yelp-chi dataset. {{Bottom: Examples of a homophilous (\cref{subfig:homophilous}) and a heterophilous (\cref{subfig:heterophilous}) region in the same graph that are both monophilous, namely they are connected to many neighbors of the same kind. In a spam network, the homophilous region corresponds to many non-spam reviews connecting to non-spam reviews (which is the expected behaviour of a non-spammer user), and the heterophilous region corresponds to spam reviews targeting non-spam reviews (which is the expected behaviour of spammers), thus, yielding a graph with both homophilous and heterophilous regions such as in \cref{subfig:distribution}.}}}
    \label{fig:homophily_distribution}
    \vspace{-1\baselineskip}
\end{figure}

\subsection{Varying Homophily} \label{sec:varying_homophily}

Graphs with monophily experience homophily, heterophily, or both. For instance, in the yelp-chi dataset -- where we classify a review as spam/non-spam (see \cref{fig:homophily_distribution}) -- we observe a case of monophily together with varying homophily. Specifically in this dataset, spam reviews are linked to non-spam reviews, and non-spam reviews usually connect to other non-spam reviews, which makes the node homophily distribution bimodal. Here the 2nd-order similarity makes the MLaP mechanism effective. 

\subsection{Theoretical Analysis} \label{sec:theoretical_analysis} 

\noindent \textbf{\emph{Justification of Stage 2:}} In Stage 2, we train a parametric model to learn the distribution of a node's neighbors from the node features $\vec \xi_i$\footnote{In \cref{sec:components}, $\vec \xi_i$s correspond to the augmented features $\vec \xi_i = [\vec x_i; \vec p_i]$}. Arguably, we can learn such a distribution na\"ively by counting the neighbors $i$ that belong to each class. This motivates our first theoretical result. In summary, we show that training a parametric model for learning the distribution of a node's neighbors (as in Stage 2) yields a lower error than the na\"ive solution. Below we present the \cref{theorem:parametric_estimation_of_q} (proof in \cref{app:proofs}) for undirected graphs (the case of directed graphs is the same, but we omit it for simplicity of exposition):

\begin{theorem} \label{theorem:parametric_estimation_of_q}
    Let $G([n], E)$ be an undirected graph of minimum degree $K > c^2$ and let $\vec Q_i \in \Gamma_c$ be the likelihood, from the viewpoint of node $i$, of any node in its neighborhood $\N(i)$ to be assigned to different classes for every node $i \in [n]$.  The following two facts are true (under standard assumptions for SGD and the losses):
    
    \begin{compactenum}
    \item Let $\vec \Qh_i$ be the sample average  of $\vec Q_i$, i.e. $\Qh_{i,j} = \frac{1}{|\N(i)|}\sum_{k \in \N(i)} \mathbb I \{ y_k = j \}$. Then, for every $i \in [n]$, we have that $\max_{j \in [c]} \E [|Q_{i,j} - \Qh_{i, j}|] \le \E [ \| \vec Q_i - \vec \Qh_i \|_\infty ] \le O \left ( \sqrt {\tfrac {\log (Kc)} {K}} \right )$.

    \item Let $q(\cdot | \vec \xi_i;  \vec \theta)$ be a model parametrized by $\vec \theta \in \R^D$ that uses the features $\vec \xi_i$ of each node $i$ to predict $\vec Q_i$. We estimate the parameter $\vec \theta_1$ by running SGD for $t = n$ steps to maximize $\L(\vec \theta) = \frac{1}{n}\sum_{i=1}^n\sum_{j=1}^c Q_{i,j}\log q(j|\vec \xi_i;\vec \theta)$. Then, for every $i \in [n]$, we have that $\max_{j \in [c]} \E [| q(j; \vec \xi_i; \vec \theta_1) - Q_{i, j} |] \le O \left ( \sqrt {\tfrac {\log n} {n}} \right )$.
    
    \end{compactenum}

\end{theorem}

It is evident here that if the minimum degree $K$ is much smaller than $n$, then the parametric model has lower error than the na\"ive approach, namely $\tilde O(n^{-1/2})$ compared to $\tilde O(K^{-1/2})$. 

\noindent \textbf{\emph{Justification of Stages 2 and 3:}} We now provide theoretical foundations for the two-stage approach. Specifically, we argue that a two-stage procedure involving learning the distribution of a node's 2nd-hop neighbor distributions (we assume for simplicity, again, that the graph is undirected) first with a parametric model such as in \cref{theorem:parametric_estimation_of_q}, and then running a two-phase algorithm to learn a parametric model that predicts a node's label, yields a lower error than na\"ively training a shallow parametric model to learn a node's labels. The first phase of the two-phase algorithm involves training the model first by minimizing the cross-entropy between the predictions and the 2nd-hop neighborhood distribution. Then the model trains a joint objective that uses the learned neighbor distributions and the actual labels starting from the model learned in the previous phase. 

\begin{theorem} \label{theorem:parametric_estimation_of_p}
    Let $G([n], E)$ be an undirected graph of minimum degree $K > c^2$ and, let $\vec P_i$ be the likelihood of node $i$ to be assigned to a different class, and let $\vec Q_i, q(\cdot | \vec \xi_i ; \vec \theta_1)$ defined as in \cref{theorem:parametric_estimation_of_q}. Let $p( \cdot | \vec \xi_i; \vec w)$ be a model parametrized by $\vec w \in \R^D$ that is used to predict the class assignments $y_i \sim p(\cdot | \vec \xi_i; \vec w)$. Let $\vec w_*$ be the optimal parameter. The following are true (under standard assumptions for SGD and the losses):
    
    \begin{compactenum}
        \item The na\"ive optimization scheme that runs SGD to maximize $\G(\vec w) = \frac{1}{n}\sum_{i=1}^n \sum_{j=1}^c P_{i,j}\log p(j|\vec \xi_i; \vec w)$ for $n$ steps has error $\E [ \G(\vec w_{n + 1}) - \G(\vec w_*) ] \le O \left ( \tfrac {\log n} {n} \right )$.

        \item The two-phase optimization scheme that runs SGD to maximize $\Gh(\vec w) = \frac{1}{n}\sum_{i=1}^n \sum_{j=1}^c {\big(\frac{1}{|\N(i)|}\sum_{k\in \N(i)} q(j|\vec \xi_k; \vec \theta_1)\big)} \log p(j|\vec \xi_i; \vec w)$ for $n_1$ steps, to estimate a solution $\vec w'$ and then runs SGD on the objective $\lambda \Gh(\vec w) + (1 - \lambda) \G(\vec w)$ for $n$ steps starting from $\vec w'$, achieves error $\E[\G(\vec w_{n+1}) - \G(\vec w_*)] \leq O\left(\frac {\sqrt{\log n \log\log n}} n \right)$.
        
    \end{compactenum}
    
\end{theorem}

You can find the proof in \cref{app:proofs}. We observe that the two-phase optimization scheme can reduce the error by a factor of $\sqrt {\log n / \log \log n}$ highlighting the importance of using the distribution of the 2nd-hop neighbors of a node to predict its label {and holds regardless of the homophily properties of the graph}. Also, note that the above two-phase optimization scheme differs from the description of the method we gave in \cref{alg:glinkx}. The difference is that the distribution of a node's neighbors is embedded into the model in the case of \cref{alg:glinkx}, and the distribution of a node's neighbors are embedded into the loss function in \cref{theorem:parametric_estimation_of_p} as a regularizer. In \cref{alg:glinkx}, we chose to incorporate this information in the model because using multiple losses harms scalability and makes training harder in practice. In the same spirit, the conception of GCNs \citep{kipf2016semi} replaces explicit regularization with the graph Laplacian with the topology into the model (see also \citep{hamilton2017inductive, yang2016revisiting}).

\subsection{Complementarity} \label{sec:complementarity}

Different components of GLINKX provide a \emph{complementary} signal to components proposed in the GNN literature \citep{maurya2021improving, zhang2022graph, rossi2020sign}. One can combine GLINKX with existing architectures (e.g. feature propagations \citep{maurya2021improving, rossi2020sign}, label propagations \citep{zhang2022graph}) for potential metric gains. For example, SIGN computes a series of $r \in \mathbb N$ feature propagations $[X, \Phi X, \Phi^2 X, \dots, \Phi^r X]$ where $\Phi$ is a matrix (e.g., normalized adjacency or normalized Laplacian) as a preprocessing step. We can include this complementary signal, namely, embed each of the propagated features and combine them in the 3rd Stage to GLINKX. Overall, although in this paper we want to keep GLINKX simple to highlight its main components, we conjecture that adding more components to GLINKX would improve its performance on datasets with highly variable homophily (see \cref{sec:varying_homophily}).

\section{Experiments \& Conclusions} \label{sec:experiments}

\noindent \textbf{\emph{Comparisons:}} We experiment with homophilous and heterophilous datasets (see \cref{tab:results} and \cref{app:datasets}). We train KGEs with Pytorch-Biggraph \citep{lerer2019pytorch, yang2014embedding}. For homophilous datasets, we compare with vanilla GCN and GAT, FSGNN, and Label Propagation (LP). For a fair comparison, we compare with one-layer GCN/GAT/FSGNN/LP since our method is one-hop. We also compare with higher-order (h.o.) GCN/GAT/FSGNN/LP with 2 and 3 layers. In the heterophilous case, we compare with LINKX\footnote{We have run our method with hyperparameter space that is a subset of the sweeps reported in \citep{lim2021large} due to resource constraints. A bigger hyperparameter search would improve our results.}
because it is scalable and is shown to work better than other baselines (e.g., H2GCN, MixHop, etc.) and with FSGNN. Note that we do not compare GLINKX with other more complex methods because GLINKX is complementary to them (see \cref{sec:complementarity}), and we can incorporate these designs into GLINKX. We use a \emph{ResNet} module to combine our algorithm's components from Stages 2 and 3. Details about the hyperparameters we use are in 
 \cref{app:hyperparameters}.

In the heterophilous datasets, GLINKX outperforms LINKX (except for arxiv-year, where we are within the confidence interval).
Moreover, the performance gap between using KGEs and adjacency embeddings shrinks as the dataset grows. In the homophilous datasets, GLINKX outperforms 1-layer GCN/GAT/LP/FSGNN and LINKX. In PubMed, GLINKX beats h.o. GCN/GAT and in arxiv-year GLINKX is very close to the performance of GCN/GAT. 

Finally, we note that our method produces consistent results across regime shifts. In detail, in the heterophilous regime, our method performs on par with LINKX; however, when we shift to the homophilous regime, LINKX's performance drops, whereas our method's performance continues to be high. Similarly, while FSGNN performs similarly to GLINKX on the homophilous datasets, we observe a significant performance drop on the heterophilous datasets (see arxiv-year). 

\begin{table*}[t]
    \centering
    \scriptsize
    \caption{Experimental results. (*) = results from the OGB leaderboard.}
    \begin{tabular}{lllHlll}
    \toprule
    
    & \multicolumn{3}{c}{Homophilous Datasets} & \multicolumn{3}{c}{Heterophilous Datasets} \\
    \midrule
    & PubMed & ogbn-arxiv &  ogbn-products & squirrel & yelp-chi & arxiv-year \\
    \midrule
    $n$ & 19.7K & 169.3K & & 5.2K & 169.3K & 45.9K \\
    $m$ & 44.3K & 1.16M & & 216.9K & 7.73M & 1.16M \\
    Edge-insensitive homophily \citep{lim2021large} & 0.66 & 0.41 & & 0.02 & 0.05 & 0.27 \\
    $d_X / c$ & 500 / 27 & 128 / 40 & &  2089 / 5 & 32 / 2 & 128 / 5 \\
    \midrule
    GLINKX w/ KGEs & {87.95 \pmt 0.30} & {\textbf{69.27 \pmt 0.25}} & 78.15 \pmt 0.10 & 45.83 \pmt 2.89 & 87.82 \pmt 0.20 & \textbf{54.09 \pmt 0.61} \\
    GLINKX w/ Adjacency & \textbf{88.03 \pmt 0.30} & 69.09 \pmt 0.13 &   & {\textbf{69.15 \pmt 1.87}} & {\textbf{89.32 \pmt 0.45}} & 53.07 \pmt 0.29 \\ 
    \midrule
    Label Propagation (1-hop) & 83.02 \pmt 0.35 &\textbf{69.59 \pmt 0.00} & & 32.22 \pmt 1.45 & 85.98 \pmt 0.28 & 43.71 \pmt 0.22 \\
 
    LINKX (from \citep{lim2021large}) & 87.86 \pmt 0.77 & 67.32 \pmt 0.24 & 69.02 \pmt 0.61 & 61.81 \pmt 1.80 & 85.86 \pmt 0.40 & {\textbf{56.00 \pmt 1.34}} \\
    LINKX (our runs) & 87.55 \pmt 0.37 & 63.91 \pmt 0.18 & & 61.46 \pmt 1.60 & 88.25 \pmt 0.24 & 53.78 \pmt 0.06 \\
    GCN w/ 1 Layer & 86.43 \pmt 0.74 & 50.76 \pmt 0.20 & 66.28 & \multicolumn{3}{c}{N/A} \\ 
    GAT w/ 1 Layer & 86.41 \pmt 0.53 & 54.42 \pmt 0.10 & 65.31 & \multicolumn{3}{c}{N/A} \\
    FSGNN w/ 1 Layer & 88.93 \pmt 0.31 & 61.82 \pmt 0.84 & 70.44 \pmt 0.15 & 64.06 \pmt 2.69 & 86.36 \pmt 0.36 & 42.86 \pmt 0.22 \\ 
    \midrule
    Higher-order GCN &  86.29 \pmt 0.46 & 71.18 \pmt 0.27 (*) & 75.64 \pmt 0.21 (*) & \multicolumn{3}{c}{N/A} \\
    Higher-order GAT  & 86.64 \pmt 0.40 & \textbf{73.66 \pmt 0.11} (*)  & 79.45 \pmt 0.59 (*) & \multicolumn{3}{c}{N/A} \\
    Higher-order FSGNN & \textbf{89.37 \pmt 0.49} & 69.26 \pmt 0.36 & 76.03 \pmt 0.33 & 68.04 \pmt 2.19 & 86.33 \pmt 0.30  & 44.89 \pmt 0.29\\
    Label Propagation (2-hop)  & 83.44 \pmt 0.35 & 69.78 \pmt 0.00 & & 43.41 \pmt 1.44 & 85.95 \pmt 0.26 & 46.30 \pmt 0.27 \\ 
    Label Prop. on $\mathbb I[A^2\!-\!A\!-\!I\!\ge\!0]$ & 82.14 \pmt 0.33 & 9.87 \pmt 0.00 & & 24.43 \pmt 1.18 & 85.68 \pmt 0.32 & 23.08 \pmt 0.13 \\
    \bottomrule
    \end{tabular}

    \vspace{-\baselineskip}
    \label{tab:results}
\end{table*}

\begin{table*}[t]
    \centering
    \scriptsize
    \caption{Ablation Study. We use the hyperparameters of the best run from \cref{tab:results} with KGEs.}
    \begin{tabular}{lllllll}
    \toprule
        & Ablation Type & Stages & All & Remove ego embeddings & Remove propagation & Remove PEs \\
    \midrule
        Heterophilous & arxiv-year & All Stages & {54.09 \pmt 0.61} & 53.52 \pmt 0.77 & 50.83 \pmt 0.24 & {39.06 \pmt 0.35} \\
        & arxiv-year & 3rd Stage & 54.09 \pmt 0.61 & 53.69 \pmt 0.65 & 50.83 \pmt 0.24 & 49.13 \pmt 1.10 \\
    \midrule
        Homophilous & ogbn-arxiv & All Stages & {69.27 \pmt 0.25} & {61.26 \pmt 0.33} & 62.70 \pmt 0.34 & 65.64 \pmt 0.18 \\
        & ogbn-arxiv & 3rd Stage & 69.27 \pmt 0.25 & 67.60 \pmt 0.39 & 62.70 \pmt 0.34 & 69.62 \pmt 0.15 \\
    \bottomrule
    \end{tabular}

    \vspace{-\baselineskip}
    \label{tab:ablation_study}
\end{table*}

\noindent \textbf{\emph{Ablation Study:}} We ablate each component of \cref{alg:glinkx} to see each component's performance contribution. We use the hyperparameters of the best model from \cref{tab:results}. We perform two types of ablations: (i) we remove each of the components from all stages of the training, and (ii) we remove the corresponding components only from the 3rd Stage. 
Except for removing the PEs from the 3rd Stage only on ogbn-arxiv, all components contribute to increased performance on both datasets. Note that adding PEs in the 1st Stage does improve performance, suggesting the primary use case of PEs. 

\noindent \textbf{\emph{Conclusion:}} We present GLINKX, a scalable method for node classification in homophilous and heterophilous graphs that combines three components: (i) ego embeddings, (ii) PEs, and (iii) monophilous propagations. As future work, (i) GLINKX can be extended in heterogeneous graphs, (ii) use more expressive methods such as attention or Wasserstein barycenters \citep{pmlr-v32-cuturi14} for averaging the low-dimensional messages, and (iii) add complementary signals.

\subsection*{Acknowledgements}

The authors would like to thank Maria Gorinova, Fabrizio Frasca, Sophie Hilgard, Ben Chamberlain, and Katarzyna Janocha from Twitter Cortex Applied Research for the useful discussions and comments on our work. MP thanks Jon Kleinberg for providing travel support through the Simons Investigator Grant. MP would also like to thank Felix Hohne for his help regarding questions about the LINKX codebase.

\newpage

\bibliography{references}
\bibliographystyle{iclr2023_conference}

\newpage 

\appendix

\begin{center}
    \Huge \textbf{Supplementary Material}
\end{center}

\section{Model Addendum} \label{app:model}

\subsection{Graph Mini-batching \& Shallow Methods} 

To train a GCN-based model (or generally, whenever propagations based on the graph topology are involved in the model) on a large network (that cannot fit in the GPU memory), one has to do \emph{minibatching} through \emph{neighbor sampling}. 
For large-scale networks, mini-batching takes much longer than full-batch training, which is one of the reasons that graph GCNs are not preferred in real-world settings (see, e.g., \citep{jin2022graph, zhang2022graphless, zheng2022cold, lim2021large, maurya2021improving, rossi2020sign}). 

For this reason, most methods that can scale on real-world settings are \emph{shallow}. \emph{Shallow} (or node-level) models are based on manipulating the node features $\vec X$ and the graph topology $\vec A$ so that propagations do not occur during training. Examples of these methods are LINKX, FSGNN \citep{maurya2021improving}, and SIGN \citep{rossi2020sign}. 
Such methods treat the input embeddings as tabular data and pass them through a feed-forward neural network (MLP) to produce the predictions. Thus, they avoid the need for neighborhood sampling and instead rely on simple tabular mini-batching. 

\subsection{Relationship to Label Propagation} 

From a label propagation perspective, one could argue that our method is related to the HITS algorithm of \citet{kleinberg1999authoritative}. Besides, on a directed graph, a potential solution would be to perform label propagation with the Gram matrix $AA^T$ and use this information as the propagated labels instead of $\vec y_i'$. This, however, would not be inductive.

\subsection{Scalability} \label{sec:scalability}

\begin{table}[H]
    \centering
    \caption{Inference Complexity (on the whole dataset) for various methods. The adjacency matrix $\vec A$ is given in sparse (CSR) format.}
    \scriptsize
    \begin{tabular}{ll}
    \toprule
        Method & Inference Complexity  \\
    \midrule
        GLINKX & $O ( mc + n (d_Xh + h^2 L_X + d_P h + h^2 L_P + h^2 L_{\text{agg}} ) )$ \\
        LINKX & $O (md + n (d_Xh + h^2 L_X + h^2 L_{\text{agg}} )$ \\
        $L$-layer GCN & $O(mdL + nd^2 L)$ \\
        $L$-hop SIGN/FSGNN & $O(nL(nd + d^2 h + L_{\text{agg}} h^2))$ \\ 
        $L$-hop $\tau$-iteration Label Propagation & $O((n + m)cL\tau)$ \\
    \bottomrule
    \end{tabular}
    \label{tab:complexity}
    \vspace{-\baselineskip}
\end{table}

\noindent \textbf{\emph{Cost of 1st Stage:}} For the 1st Stage, we are using Pytorch-Biggraph to train PEs such as TransE \citep{bordes2013translating}, DistMult \citep{yang2014embedding}. Please refer to \citep{lerer2019pytorch} for more information. 

\noindent \textbf{\emph{Cost of 2st Stage:}} For the 2nd Stage we pay \textbf{once} $O(m_{\text{train}} c)$ ($m_{\text{train}} = |E_{\text{train}}|$ is the number of edges in the graph induced by the train nodes) to do perform MLaP Forward and then the MLP that takes $\vec x_i$ and $\vec p_i$ with $L_X$ layers for the features $\vec x_i$ and $L_P$ layers for the PEs $\vec p_i$, and $L_{\text{agg}}$ layers for the ResNet and hidden dimension $h$, has forward pass complexity $O\left (n (d_Xh + h^2 L_X + d_P h + h^2 L_P + h^2 L_{\text{agg}} ) \right )$. Finally, MLaP Backward costs $O(m_{\text{train}}c)$, similarly to MLaP Forward.

\noindent \textbf{\emph{Cost of 3rd Stage:}} The final model costs $O (n (d_Xh + h^2 L_X + d_P h + h^2 L_P + ch + h^2 L_{\text{prop}} + h^2 L_{\text{agg}}  ) )$ for a forward pass of the MLP, where $c$ is the class dimensionality and $L_{\text{prop}}$ are the layers for the MLP handling the propagated labels $\vec y_i'$.

\noindent \textbf{\emph{Inference Complexity:}} \cref{tab:complexity} shows the inference complexity of GLINKX and compares it with the complexity of other methods such as LINKX, $L$-layer GCN, $L$-hop FSGNN/SIGN and LP. 

Compared to LINKX we pay an $O (m_{\text{train}} c)$ extra cost once. Regarding embedding the adjacency matrix LINKX pays a $O(mh + n h^2 L_P)$ cost for generating the $h$-dimensional adjacency embeddings whereas we pay $O(n d_P h + n h^2 L_P)$ which is better for $d_P = O(m / n)$. This holds in most real-world large networks since $d_P \sim 10^2$ whereas $n \sim 1 \mathrm B$ and $m \sim 100 \mathrm B$. Also note that using KGEs as PEs has an additional benefit, since KGEs can be trained only \emph{once} and can be used off-the-shelf for other downstream tasks. An $L$-layer GNN propagates using the adjacency matrix $\vec A$ and multiplies with a projection matrix $\vec W^{(i)}$ (for each layer) thus achieving an $O(dLm + n d^2 L )$ complexity; the former term is due to the propagation (assuming $\vec A$ is sparse), and the latter term accounts for the multiplication with the projection matrix. The $L$-hop SIGN (resp. $L$-hop FSGNN) performs $L$ (resp. $2L$) propagations of the feature matrix $\vec X$ using a sequence of matrices $\{ \Phi^\ell \}_{\ell \in [L]}$ which cost a total of $O(n^2 d L)$ (assuming the matrices are dense), and then uses a ResNet with $L_{\text{agg}}$ layers block to concatenate the propagated features, which costs $O(nLdhL_{\text{agg}})$, yielding an inference complexity cost of $O(nL(nd + d^2 h + L_{\text{agg}} h^2))$. Finally, Label Propagation with sparse matrices costs $O((n + m) c L \tau)$, where $\tau$ is the number of iterations of the LP algorithm.

\subsection{Knowledge Graph Embedding Training} \label{sec:kge_training}

{In the sequel, we provide a general algorithm for training KGEs to be used as positional embedding $\vec P$. Briefly we treat $G$ as a knowledge graph with only one relation $\vec r = \vec \rho$. Specifically, each edge $(u, v) \in E$ corresponds to a head entity with embedding $\vec h = \vec p_u$ and a tail entity with embedding $\vec t = \vec p_v$. Without loss of generality we can pick $\vec \rho = \one$ to be the vector of all 1s since the constants per dimension can be absorbed by the positional embeddings.} 

\begin{algorithm}[H]
    \caption{Knowledge Graph Embedding Training} \label{alg:kge_training}
    \begin{flushleft}
        \textbf{Input:} Graph $G(V, E)$, Embedding dimension $d_P$, Comparison function $f : \mathbb R^{d_P} \times \mathbb R^{d_P} \times \mathbb R^{d_P} \to \mathbb R$, Sample loss function $\ell : \mathbb R \times \mathbb R \to \mathbb R$, Number of epochs $T$, Batch Size $B$, Number of negative samples $\nu$ \\
        \textbf{Output:} Positional Embeddings $\vec P \in \mathbb R^{n \times d_P}$
        
        Randomly, initialize the entity embedding matrix $\vec P \in \mathbb R^{n \times d_P}$ with row vectors $\vec p_u$ for all $u \in V$. Let $\vec \rho = \one$ \\
    
        For each epoch $t \in [T]$, sample minibatches $\{ E_{\text{batch}} \}$ of the edges $E$ of size $B$. 
        
        \begin{enumerate}
            \item For each minibatch $E_{\text{batch}}$ and each $(u, v) \in E$ sample $\nu$ negative samples $(u_i', v_i') \notin E$ and build $T_{\text{batch}} = \bigcup_{i \in [\nu]} \{ ((u, v), (u_i', v_i')) \} $
            \item Update the embeddings $\vec P$ wrt 
            
            \begin{align*}
                \nabla_{\vec P} \sum_{((u, v), (u', v')) \in T_{\text{batch}}} \ell(f(\vec p_u, \vec \rho, \vec p_v), f(\vec p_{u}', \vec \rho, \vec p_{v}'))
            \end{align*}
            
        \end{enumerate}
    
    Output $\vec P$
    \end{flushleft}

\end{algorithm}

{In our case, we have chosen the DistMult embeddings \cite{yang2014embedding} which correspond to choosing the triple product $$f(\vec h, \vec r, \vec t) = \sum_{i \in d_P} h_i r_i t_i$$ as a comparison function, and the margin-based ranking loss $\ell(f(\vec h, \vec r, \vec t), f(\vec h', \vec r, \vec t')) = \max \{ 0, \gamma + f(\vec h, \vec r, \vec t) - f(\vec h', \vec r, \vec t')\}$ where $\gamma$ is the margin.}

\section{Inductive GLINKX} \label{app:inductive}

In the paper we focus on the transductive setting. Here we show how we can extend our framework to the inductive setting. In the inductive setting, during training we only have access to the training graph $G_{\text{train}}(V_{\text{train}}, E_{\text{train}})$. During test, the whole graph $G \subseteq G_{\text{train}}$ is revealed and we make predictions for the test set. The stages of GLINKX in the inductive setting are as follows 

\noindent \textbf{1st Stage.} For the KGE pretraining Stage, we train KGEs on $G_{\text{train}}$. Then we can use existing methods in the literature such as \citep{el2022twhin, albooyeh2020out} to infer the KGEs of the test nodes.
   
\noindent \textbf{2nd Stage.} We train the model that predicts the distribution of the neighbors on $G_{\text{train}}$ as in \cref{alg:glinkx}. Then for the test nodes, we know their ego features $\vec x_i$ and we can also infer their PEs (see above) from the pre-trained PEs on the training set. We use these and the pre-trained shallow model in order to predict $\vec {\tilde y}_i$ for the test nodes. Then, we run MLaP backward to compute $\vec {y}'_i$. 

\noindent \textbf{3rd Stage.} First, we train the second shallow model by using the propagated soft-predictions only on $G_{\text{train}}$. We then use the ego features, PEs and the propagated information $\vec {y}_i'$ to make predictions on the test set. 

\cref{alg:glinkx_inductive} shows the process of making predictions on the test set. 

\begin{algorithm}[h]

\caption{Inductive GLINKX} 
\label{alg:glinkx_inductive}
\begin{flushleft}
\textbf{Input:} Graph $G(V, E)$ with train set $V_{\text{train}} \subseteq V$ and test set $V_{\text{test}} \subseteq V$, node features $X$, labels $Y$ \\
\textbf{Output:} Predictions for all nodes $i \in V_{\text{test}}$\\

\textbf{Pre-training.} Call \cref{alg:glinkx} with input graph $G_{\text{train}}(V_{\text{train}}, E_{\text{train}})$ and compute PEs $\{ \vec p_i \}_{i \in V_{\text{train}}}$, and pre-trained models $f_1$ (from 2nd Stage) and $f_2$ (from 3rd Stage) 

\medskip

\textbf{1st Stage.} Create PEs $\{ \vec p_i \}_{i \in V_{\text{test}}}$ for the nodes of the test set (see e.g. \citep{el2022twhin, albooyeh2020out})

\medskip

\textbf{2nd Stage.} Predict the distribution of neighbors for the nodes of the test set as $\vec {\tilde y}_i = f_1(\vec x_i, \vec p_i)$ for all $i \in V_{\text{test}}$. Calculate $\vec {y}_i' = \frac {\sum_{j \in V : (i, j) \in E} \vec {\tilde y}_j} {| \{ j \in V : (i, j) \in E \}|}$ for all $i \in V_{\text{test}}$ (MLaP Backward).

\medskip

\textbf{3rd Stage.}  Compute $\vec y_{\text{final}, i} = f_2(\vec x_i, \vec p_i, \vec {y}_i')$ for all $i \in V_{\text{test}}$. 

\medskip

Return $\{ \vec y_{\text{final}, i} \}_{i \in V_{\text{test}}}$
\end{flushleft}
\end{algorithm}






\section{Hyperparameters} \label{app:hyperparameters}

For each of the methods we train for 200 epochs and report the test set accuracy on the epoch with the best validation accuracy. 

\subsection{KGEs} \label{app:biggraph_hyperparameters}

We use the following hyperparameters for the KGEs using Pytorch-Biggraph: 

\begin{compactitem}
    \item dimension = 400
    \item 50 epochs
    \item negative samples = 1000
    \item batch size = 10000
    \item \texttt{dot} comparator, \texttt{softmax} loss \citep{yang2014embedding}
    \item learning rate = 0.1
\end{compactitem}

\subsection{GLINKX w/ Adjacency} \label{app:glinkx_hyperparameters}

\subsubsection{Sweeps}

We perform the following sweeps: 

\begin{itemize}
	\item glinkx\_init\_layers\_X $\in \{ 1, 2 \}$
	\item glinkx\_init\_layers\_A"$\in \{ 1, 2 \}$ 
	\item glinkx\_init\_layers\_agg $\in \{ 1, 2 \}$
	\item glinkx\_inner\_dropout : $\in \{ 0.5 \}$
    \item lr $\in \{ 0.1, 0.001 \}$
	\item optimize:r: AdamW
\end{itemize}

\input{glinkx_fast_2_hyperparameters.tex}

\subsection{GLINKX w/ KGEs} \label{app:glinkx_hyperparameters}

\subsubsection{Sweeps}

We perform the following sweeps for all datasets 

\begin{itemize}
	\item glinkx\_init\_layers\_X $\in \{ 1, 2 \}$
	\item glinkx\_init\_layers\_A $\in \{ 1, 2 \}$ 
	\item glinkx\_init\_layers\_agg $\in \{ 1, 2 \}$
	\item glinkx\_inner\_dropout : $\in \{ 0.5 \}$
    \item lr $\in \{ 0.1, 0.001 \}$
	\item optimize:r: AdamW
	\item biggraph\_vector\_length: 400
\end{itemize}

\input{glinkx_fast_2_biggraph_hyperparameters.tex}

\subsection{GAT} \label{app:gat_hyperparameters} 

\subsubsection{Sweeps}

\begin{itemize}
	\item gat\_num\_layers $\in \{ 1 \}$
	\item gat\_hidden\_channels $\in \{ 4, 8, 16, 32 \}$
	\item gat\_num\_heads $\in \{ 2, 4 \}$
	\item lr $\in \{ 0.01, 0.001 \}$
\end{itemize}

\input{gat_hyperparameters.tex}

\subsection{GCN} \label{app:gcn_hyperparameters}

\subsubsection{Sweeps} 

We perform the following sweeps: 

\begin{itemize}
	\item gcn\_num\_layers $\in \{ 1 \}$
	\item gcn\_hidden\_channels $\in \{ 4, 8, 16, 32, 64 \}$
	\item lr $\in \{ 0.01, 0.001 \}$
\end{itemize}

\input{gcn_hyperparameters.tex} 

\subsection{FSGNN}

We run the following sweeps for FSGNN

\begin{itemize}
    \item layers = 1 for the 1-layer case and layers $\in \{ 2, 3 \}$ for the higher-order case \item hidden channels $\in \{ 32, 64, 128 \}$
    \item learning rate $\in \{ 0.01, 0.001 \}$
    \item layer normalization $\in \{ \text{true}, \text{false} \}$
\end{itemize}

\subsection{Label Propagation}

We run the following sweeps using the implementation of Label Propagation from \citep{lim2021large}: 

\begin{itemize}
    \item $\alpha \in \{ 0.01, 0.1, 0.25, 0.5, 0.75, 0.99 \}$
    \item $\text{hops } \in \{ 1, 2 \}$
\end{itemize}

\section{Experiments Addendum} \label{app:experiments_addendum}

\subsection{Implementation and Environment} \label{app:implementation_environment} 

GLINKX is implemented in Pytorch-Geometric. For the knowledge graph embeddings we use the official Pytorch-Biggraph implementation. We use the official implementation of LINK.
For hardware we used Vertex AI notebooks with 8 NVIDIA Tesla V100 with 16GB of memory and 96 CPUs. 

\subsection{Experimental Protocol} \label{app:experimental_protocol}

\noindent \textbf{Error Bars.} For the PubMed dataset and the heterophilous datasets provided by \citep{pei2020geom, lim2021large, yang2016revisiting} we used the fixed splits provided with a fixed seed. For the OGB datasets, where there exists only one officialy released split for each dataset \citep{hu2020open} we generate the error bars by runnning the respective experiments with 10 different seeds (0-9).

\subsection{Datasets} \label{app:datasets}

For our experiments we use the following datasets (see \cref{tab:results} for the dataset statistics): 

\begin{compactitem}
    \item \emph{PubMed \citep{yang2016revisiting,sen2008collective}.} The PubMed dataset consists of scientific publications from PubMed database pertaining to diabetes classified into one of three classes.  Each node is described by a TF-IDF weighted word vector from a dictionary which consists of 500 unique words. 
    
    \item \emph{ogbn-arxiv \citep{hu2020open,Bhatia16,sinha2015overview}.} The ogbn-arxiv dataset is a directed graph, representing the citation network between all CS papers on arxiv mined from the Microsoft Academic Graph (MAG). Nodes are papers and edges correspond to citations. The node features correspond to the average of word embeddings of the title and abstract of the papers. The task is to predict the papers' subcategories (e.g. CS.LG, CS.SI etc.). 
    
    \item \emph{squirrel \citep{rozemberczki2021multi}.} The data represents page-to-page networks on squirrels mined from Wikipedia between October 2017 and November 2018. Node represent articles and edges are mutual links between the pages. The features of each node are binary and represent the existence of informative nouns that appear on the corresponding Wikipedia pages. 
    
    \item \emph{arxiv-year, yelp-chi \citep{lim2021large}.} The arXiv-year dataset is derived from the ogbn-arxiv dataset where the labels have been changed in order to convert the dataset from homophilous to heterophilous. Instead of predicting each paper's subcategories, the new task focuses on predicting the year that a paper is posted, where batches of years have been converted to labels. The yelp-chi dataset includes hotel and restaurant reviews filtered (spam) and recommended (legitimate) by Yelp. The graph structure comes from \citep{dou2020enhancing} and the features from \citep{sen2008collective}. The task is to predict whether a review is a spam or not. 
    
\end{compactitem}

\section{Stages Diagram}

\begin{figure}[H]
    \centering
    \includegraphics[width=0.8\columnwidth]{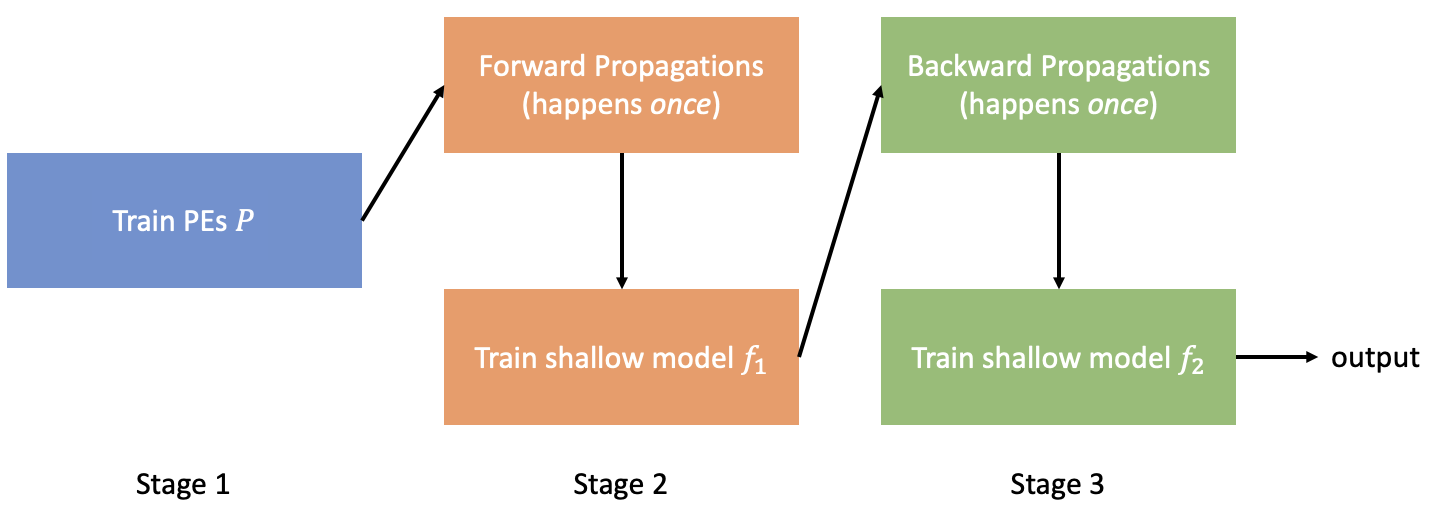}
    \caption{GLINKX stages.}
    \label{fig:stages}
\end{figure}

\section{Proofs} \label{app:proofs}

\subsection{Proof of \cref{theorem:parametric_estimation_of_q} (Parametric Estimation of $Q$)}

\textbf{Counting Estimator.} As the first step, we can estimate $Q_i$, denoted by $\Qh_i$ by taking the average class assignment in its neighbor $\N(i)$, i.e.
\[
    \Qh_{i,j} = \frac{1}{|\N(i)|}\sum_{k \in \N(i)} \mathbb I \{ y_k = j \}
\]
Evidently, $\E[\Qh_{i,j}] = Q_{i,j}$. Also, using Hoeffiding concentration inequality, we have, with a probability at least $1 - \delta$, 
\[
    \left|\Qh_{i,j} - \frac{1}{|\N(i)|}\sum_{k \in \N(i)} P_{k,j}\right| \leq  \sqrt{\frac{1}{2K}\log\frac{2}{\delta}}
\]
Since 
\[
\vec Q_i = \frac{1}{|\N(i)|}\sum_{k \in \N(i)} \vec P_k
\]
we have 
\[
    \|\vec \Qh_i - \vec Q_i\|_{\infty} \leq \sqrt{\frac{1}{2K}\log\frac{2c}{\delta}}
\]

with probability at least $1 - \delta$. Therefore, by the law of total expectation and by minimizing wrt $\delta$, we get 

\begin{eqnarray*}
\max_{j \in [c]} \E [ | Q_{ij} - \Qh_{ij} | ] \le \E [ \| \vec Q_i - \vec \Qh_i \|_\infty ] \le \min_{\delta \in (0, 1)} \sqrt{\frac{1}{2K}\log\frac{2c}{\delta}} + \delta \le O \left ( \sqrt {\frac {\log (Kc)} {K}} \right )
\end{eqnarray*} 

\textbf{Parametric Estimator.} As the first step, we show that through a parametric optimization, we will be able to obtain a better estimation of $\vec Q_i$. In particular, we assume that all $\vec Q_i$ can be written in the parametric form $q(j|\vec \xi_i;\vec \theta)$, where $\vec \theta \in \R^D$ is the parameter of the model. Let $\vec \theta_*$ be the optimal model parameter such that $Q_{i,j} = q(j|\vec \xi_i;\vec \theta_*)$. We define the following loss function
\[
\L(\vec \theta) = -\frac{1}{n}\sum_{i=1}^n\sum_{j=1}^c Q_{i,j}\log q(j|\vec \xi_i;\vec \theta)
\]
Evidently, $\vec \theta_*$ is the minimizer of $\L(\vec \theta)$. We assume that $\L(\vec \theta)$ is $L$ smooth and $\gamma$-PL function, i.e.
\[
\| \nabla \L (\vec \theta) \| \le L, \quad \L(\vec \theta) - \L(\vec \theta_*) \leq \frac{1}{2\gamma}\|\nabla\L(\vec \theta)\|^2
\]
The real objective function $\Lh(\vec \theta)$ based on the observation of class assignments is given as
\[
\Lh(\vec \theta) = -\frac{1}{n}\sum_{i=1}^n\sum_{j=1}^c \Qh_{i,j}\log q(j|\vec \xi_i;\vec \theta)
\]
To find the optimal $\vec \theta$, we run stochastic gradient descent. At each iteration $t$, we sample one node $i_t$, and compute the gradient $\vec g_t$ as
\[
    \vec g_t = \sum_{j=1}^c \Qh_{i,j}\nabla_{\vec \theta} \log q(j|x_{i_t};\vec \theta)
\]
and update the solution by
\[
    \vec \theta_{t+1} = \vec \theta_{t} - \eta \vec g_t
\]
where $\eta$ is a fixed step size whose value will be decided later. We have
\begin{eqnarray*}
\lefteqn{\E[\L(\vec \theta_{t+1}) - \L(\vec \theta_t)]} \\
& \leq & \E\left[\langle \nabla \L(\vec \theta_t), \vec \theta_{t+1} - \vec \theta_t\rangle\right] + \frac{L}{2}\E\left[\|\vec \theta_{t+1} - \vec \theta_t \|^2\right] \\
& = & -\eta\left(1 - \frac{\eta L}{2}\right)\E\left[\|\nabla \L(\vec \theta_t)\|^2\right] + \frac{\eta^2L}{2}\E\left[\|\vec g_t - \vec g_t'\|^2 + \|\vec g_t' - \nabla \L(\vec \theta_t)\|^2\right]
\end{eqnarray*}
where $\vec g_t' = \sum_{j=1}^c Q_{i,j} \nabla_{\vec \theta} \log q(j|x_{i_t};\vec \theta)$. We assume that $\|\nabla_{\vec \theta}\log q(j|\vec \xi_i;\vec \theta)\| \leq G$ for any $\vec \theta$ and any $i$. We have
\[
    \E\left[\|\vec g_t - \vec g_t'\|^2\right] \leq \left(\frac{c^2}{K}\log\frac{2}{\delta} + \delta\right)G^2, \quad \E\left[\|\vec g_t' - \nabla \L(\vec \theta_t)\|^2\right] \leq G^2
\]
By choosing $\delta = c^2/K$, we have
\[
\E\left[\|\vec g_t - \vec g_t'\|^2\right] \leq \frac{2c^2 G^2}{K}\log\frac{2K}{c^2}
\]
Putting the above bounds together, we have
\[
\E\left[\L(\vec \theta_{t+1}) - \L(\vec \theta_t)\right] \leq - \eta\left(1 - \frac{\eta L}{2}\right)\E\left[\|\nabla \L(\vec \theta_t)\|^2\right] + \frac{\eta^2 G^2 L}{2}\left(1 + \frac{2c^2}{K}\log\frac{2K}{c^2}\right)
\]
Assume $\eta L \leq 1$. Using the fact
\[
    \L(\vec \theta_t) - \L(\vec \theta_*) \leq \frac{1}{2\gamma}\|\nabla \L(\vec \theta_t)\|^2
\]
we have
\[
\E\left[\L(\vec \theta_{t+1}) - \L(\vec \theta_*)\right] \leq \left(1 - \frac{\eta\gamma}{4}\right)\E\left[\L(\vec \theta_t) - \L(\vec \theta_*)\right] + \frac{\eta^2 G^2 L}{2}\left(1 + \frac{2c^2}{K}\log\frac{2K}{c^2}\right)
\]
and therefore
\[
\E\left[\L(\vec \theta_{t+1}) - \L(\vec \theta_*)\right] \leq \exp\left(-\frac{\eta \gamma t }{4}\right)\left(\L(\vec \theta_0) - \L(\vec \theta_*)\right) + \frac{\eta G^2 L}{2}\left(1 + \frac{2c^2}{K}\log\frac{2K}{c^2}\right)
\]
Let $\vec \theta_1 = \vec \theta_{n+1}$. By choosing 
\[
    \eta = \frac{4}{n\gamma}\log \left ( \frac{n\gamma(\L(\vec \theta_0) - \L(\vec \theta_*))}{2G^2L(1 + 2c^2/K\log(2K/c^2))} \right )
\]
we have
\[
\E\left[\L(\vec \theta_1) -\L(\vec \theta_*)\right] \leq \frac{2G^2 L}{n\gamma}\left(1 + \frac{2c^2}{K}\log\frac{2K}{c^2}\right)\log\frac{n\gamma(\L(\vec \theta_0) - \L(\vec \theta_*))}{2G^2L(1 + 2c^2/K\log(2K/c^2))} = O\left(\frac{\log n}{n\gamma}\right)
\]
Since
\[
\L(\vec \theta_1) - \L(\vec \theta_*) = \frac{1}{n}\sum_{i=1}^n\sum_{j=1}^c q(j|\vec \xi_i;\vec \theta_*)\log\frac{q(j|\vec \xi_i;\vec \theta_1)}{q(j|\vec \xi_i;\vec \theta_*)} = \frac{1}{n}\sum_{i=1}^n\sum_{j=1}^c\mbox{KL}\left(q(\cdot|\vec \xi_i;\vec \theta_*)\|q(\cdot|\vec \xi_i;\vec \theta_1)\right)
\]
and (by Pinkser's Inequality), 
\[
\mbox{KL}\left(q(\cdot|\vec \xi_i;\vec \theta_*)\|q(\cdot|\vec \xi_i;\vec \theta_1)\right) \geq 2\left(\sum_{j=1}^c |q(j|\vec \xi_i;\vec \theta_1) - q(j|\vec \xi_i;\vec \theta_*)|\right)^2
\]
we have
\[
\frac{1}{n}\sum_{i=1}^n \E\left[\left(\sum_{j=1}^c|q(j|\vec \xi_i;\vec \theta_1) - Q_{i,j}|\right)^2\right] \leq \frac{C\log n}{n\gamma}
\]
where 
\[
C = G^2 L\left(1 + \frac{2c^2}{K}\log\frac{2K}{c^2}\right)\log\frac{\gamma (\L(\vec \theta_0) - \L(\vec \theta_*))}{2G^2L(1 + 2c^2/K\log(2K/c^2)}
\]
Hence, on average, we have
\[
    |q(j|\vec \xi_i;\vec \theta_1) - Q_{i,j}| \leq \sqrt{\frac{C\log n}{n\gamma}}
\]
which could be significantly smaller than $O(1/\sqrt{K})$ when $n \gg K$, indicating that through the parametric modeling of $Q_{i,j}$ and optimization of $\Lh(\vec \theta)$, we will be able to achieve significantly better estimation of $Q_{i,j}$ than a simple counting. \qed

\subsection{Proof of \cref{theorem:parametric_estimation_of_p} (Parametric Estimation of $P$)}

\noindent \textbf{Na\"ive Approach.} We now will utilize the estimation $q(\cdot|\vec \xi_i;\vec \theta_1)$ further estimate $\vec P_i$. Similar to last section, we introduce a parametric estimator $p(j|\vec \xi_i;\vec w)$, where $w \in \R^d$ is the parameter. We denote by $\vec w_*$ the optimal parameter, i.e. $P_{i,j} = p(j|\vec \xi_i; \vec w_*)$. Let $\G(\vec w)$ be the objective function based on population average, i.e.
\[
\G(\vec w) = \frac{1}{n}\sum_{i=1}^n \sum_{j=1}^c P_{i,j}\log p(j|\vec \xi_i; \vec w)
\]
Evidently, $w_*$ is a minimizer of $\G(\vec w)$. Similar to the last Section, we assume $\G(\vec w)$ is $L$ smooth and $\gamma$-PL function. 

A straightforward approach for optimizing $\G(\vec w)$ is, at each iteration $t$, sample a node $i_t$ from the network, and compute the stochastic gradient as
\[
    \vec g_t = y_{i_t}\nabla_{\vec \theta}\log p(y_{i_t}|\vec \xi_{i_t};\vec \theta_t)
\]
and update the solution as
\[
    \vec w_{t+1} = \vec w_t - \eta \vec g_t
\]
Following the same analysis from the last Section, we have

$$\E \left [ \G (\vec w_{n + 1}) - \G(\vec w_*) \right ] \le O \left ( \frac {\log n} {n \gamma} \right ) $$ 

and

\[
\frac{1}{n}\sum_{i=1}^n \E\left[\sum_{j=1}^c |p(j|\vec \xi_i; \vec w_{n+1}) - P_{i,j}|^2\right] \leq O\left(\sqrt{\frac{\log n}{n\gamma}}\right)
\]

Here, we propose a different version of SGD, with the aim to reduce the impact of variance due to the random sample of class label $y_i$. We divide optimization into two phase. In the first phase, we will optimize $\vec w$ with respect to the following objective function
\[
\Gh(\vec w) = \frac{1}{n}\sum_{i=1}^n \sum_{j=1}^c \underbrace{\left(\frac{1}{|\N(i)|}\sum_{k\in \N(i)} q(j|\vec \xi_k; \vec \theta_1)\right)}_{:= \Ph_{i,j}} \log p(j|\vec \xi_i; \vec w)
\]
In the second phase, we will run the SGD optimization that mixes stochastic gradient with the gradient of $\Gh(\vec w)$, i.e. optimizes the objective $\lambda \Gh(\vec w) + (1 - \lambda) \G(\vec w)$. 

\noindent \textbf{Phase I.} In Phase I, we will update the solution by
\[
\vec w_{t+1} = \vec w_t - \eta \nabla \Gh(\vec w_t)
\]
Using the standard analysis, we have
\begin{eqnarray*}
\lefteqn{\E[\G(\vec w_{t+1}) - \G(\vec w_t)]} \\
& \leq & -\eta \E\left[\langle \nabla \G(\vec w_t), \Gh(\vec w_t)\rangle\right] + \frac{\eta^2 L}{2}\|\nabla \Gh(\vec w_t)\|^2 \\
& = & -\frac{\eta}{2}\E[|\nabla \G(\vec w_t)|^2] - \frac{\eta}{2}(1-\eta L)\E[\|\nabla \Gh(\vec w)\|^2] + \frac{\eta}{2}\E[\|\nabla \G(\vec w_t) - \nabla \Gh(\vec w_t)\|^2]
\end{eqnarray*}
Since

\begin{eqnarray*}
\lefteqn{\E \left [ \|\nabla \G(\vec w_t) - \nabla \Gh(\vec w_t)\|^2 \right ]|} \\ & = &  \E\left \|\frac{1}{n}\sum_{i=1}^n\sum_{j=1}^c(P_{i,j} - \Ph_{i,j})\nabla \log p(j|\vec \xi_i;\vec w_t)\right \|^2 \leq \frac{G^2}{n^2}\E\left|\sum_{i=1}^n \|\vec P_i - \vec \Ph_i\|_1 \right|^2 \leq \frac{CG^2\log n}{n\gamma}
\end{eqnarray*}
and by further assuming $\eta L \leq 1$, we have
\[
\E[\G(\vec w_{t+1}) - \G(\vec w_t)] \leq -\frac{\eta}{2}\E[\|\nabla \G(\vec w_t)\|^2]
+ \frac{\eta CG^2\log n}{n\gamma}
\]
or
\[
\E[\G(\vec w_{t+1}) - \G(\vec w_*)] \leq (1 - \eta\gamma)\E[\G(\vec w_t) - \G(\vec w_*)] + \frac{\eta CG^2\log n}{n\gamma}
\]
By taking the telescope sum, we have
\[
\E[\G(\vec w_{t+1} - \G(\vec w_*)] \leq \exp\left(-\gamma \eta t\right)\left(\G(\vec w_0) - \G(\vec w_*)\right) + \frac{CG^2\log n}{n\gamma}
\]
By choosing step size $\eta = 1/L$, and setting $t$ as
\[
    t = \frac{L}{\gamma}\log \frac{n\gamma(\G(\vec w_0) - \G(\vec w_*))}{CG^2\log n}
\]
we obtain the solution $w'$ with guarantee
\[
\E[\G(\vec w') - \G(\vec w_*)] \leq \frac{2CG^2\log n}{n\gamma}
\]

\noindent \textbf{Phase II.} In the second phase, we use $\vec w'$ as the initial solution for $\vec w$, and ran a SGD. At each iteration $t$, we randomly sample a node $i_t$ from the graph, compute the gradient as
\[
    \vec g_t = (1-\lambda) y_{i_t}\nabla \log p(y_{i_t}|x_{i_t};\vec w_t) + \lambda \nabla \Gh(\vec w_t)
\]
and update the solution as
\[
\vec w_{t+1} = \vec w_t - \eta \vec g_t
\]
Following the analysis from the last section, we have
\begin{eqnarray*}
\lefteqn{\E[\G(\vec w_{t+1}) - \G(\vec w)]} \\
& \leq & -\eta\E\left[\langle \nabla \G(\vec w_t), (1-\lambda)\nabla \G(\vec w_t) + \lambda \Gh(\vec w_t)\right] + \frac{\eta^2 L}{2}\E\left[\|\lambda \nabla \Gh(\vec w_t) + (1- \lambda) \G(\vec w_t) \|^2\right] + \frac{\eta^2 \eta^2 G^2 L}{2} \\
& \leq & - \frac{\eta}{2}\E\left[\|\nabla \G(\vec w_t)\|^2\right] - \frac{\eta}{2}\left(1 - \eta L\right) \E\left[\|\lambda \nabla \Gh(\vec w_t) + (1- \lambda) \nabla \G(\vec w_t)\|^2 \right] + \frac{\eta\lambda^2}{2}\E\left[\|\nabla \G(\vec w_t) - \nabla \Gh(\vec w_t)\|^2\right] \\
&  & + \frac{(1-\lambda)^2\eta^2G^2 L}{2}
\end{eqnarray*}
By choosing $\eta \leq 1/L$, we have
\[
\E\left[\G(\vec w_{t+1}) - \G(\vec w_t)\right] \leq -\frac{\eta}{2}\E\left[\|\nabla \G(\vec w_t)\|^2\right] + \frac{(1-\lambda)^2\eta^2 G^2 L}{2} + \frac{\eta\lambda^2}{2}\E\left[\|\G(\vec w_t) - \Gh(\vec w_t)\|^2\right]
\]

\begin{eqnarray*}
\lefteqn{\|\nabla \G(\vec w_t) - \nabla \Gh(\vec w_t)\|^2} \\ & = & \left\|\frac{1}{n}\sum_{i=1}^n\sum_{j=1}^c(P_{i,j} - \Ph_{i,j})\nabla \log p(j|\vec \xi_i;\vec w_t)\right\|^2 \leq \frac{G^2}{n^2}\left|\sum_{i=1}^n \| \vec P_i - \vec \Ph_i\|_1\right|^2 \leq \frac{CG^2\log n}{n\gamma}
\end{eqnarray*} 
we have
\begin{eqnarray*}
\lefteqn{\E[\G(\vec w_{t+1}) - \G(\vec w_t)]} \\
& \leq & -\frac{\eta}{2}\E[\|\nabla \G(\vec w_t)\|^2] + \frac{(1-\lambda)^2\eta^2 G^2 L}{2} + \frac{\eta CG^2\lambda^2\log n}{n\gamma}
\end{eqnarray*}
Using the standard analysis, we have
\[
\E[\G(\vec w_{t+1}) - \G(\vec w_*)] \leq \exp\left(-\eta\gamma t\right)(\G(\vec w') - \G(\vec w_*)) + \frac{CG^2\lambda^2\log n}{n\gamma} + \frac{(1-\lambda)^2\eta G^2 L}{2}
\]
By minimizing over $\lambda$, we have
\begin{eqnarray*}
\E[\G(\vec w_{t+1}) - \G(\vec w_*)] & \leq & \exp\left(-\eta\gamma t\right)(\G(\vec w') - \G(\vec w_*)) + G^2\sqrt{\frac{CL\eta\log n}{2n\gamma}} \\
& \leq & \frac{2CG^2\log n}{n\gamma} \exp(-\eta\gamma t) + G^2\sqrt{\frac{CL\eta\log n}{2n\gamma}}
\end{eqnarray*}
With $t = n$ and choosing 
\[
\eta = O\left(\frac{\log\log n}{n\gamma}\right)
\]
we have
\[
    \E[\G(\vec w_{n+1}) - \G(\vec w_*)] \leq O\left(\frac{1}{n\gamma}\sqrt{\log n \log\log n}\right)
\]
Comparing the naive approach where we simply train over class assignment $y_i$, we are able to reduce the error by a factor of $\sqrt{\log n/\log\log n}$.
\qed

\section{Further Related Work} \label{app:further_related_work} 

\textbf{Methods for Homophily and Heterophily.} 
Many methods have been adapted to work both in homophilous and heterophilous regimes. H2GCN \citep{zhu2020beyond} is one of the first methods shown to work in both kinds of datasets. RAW-GNN \citep{jin2022raw} is a random-walk-based GCN that exploits both homophily and heterophily by doing random walks and aggregations in two ways: breadth-first for homophily and depth-first for heterophily. CPGNN \citep{zhu2021graph} is a GCN-based architecture that uses a compatibility matrix for modeling the heterophily or homophily level in the graph, which can be learned in an end-to-end fashion, enabling it to go beyond the assumption of strong homophily. GPR-GNN \citep{chien2020adaptive} addresses feature over-smoothing and homophily/heterophily by combining GNNs with Generalized PageRank techniques, where each step of feature propagation is associated with a learnable weight. The amplitudes of the weights trade off the degree of smoothing of node features and the aggregation power of topological features. However, all the above methods suffer from the same scalability issues that GCNs have. 

Regarding scalable methods for both homophily and heterophily, the FSGNN \citep{maurya2021improving} method is closely related to our work. It relies on using propagations that are separate from the training (similar to SIGN) and uses separate (higher-order) feature propagations for homophily and heterophily. In contrast, our one-hop method relies on different components to address homophily and heterophily. Furthermore, FSGNN components can be added to our method, together with the ego features, the PEs, and the propagations, and thus their work can also be seen as complementary to ours. CLP \citep{zhong2022simplifying} combines a shallow (base) predictor model with a modified Label Propagation that works both in homophily and heterophily by using a class compatibility matrix (similarly to CPGNN) for the Label Propagation step. Our method is fundamentally different; however, incorporating a compatibility matrix for the propagation stage constitutes interesting future work. 

\textbf{Positional Embeddings.} Following up on the recent success of LINKX in classifying nodes in heterophilous settings based partially on the \emph{position} of each node (adjacency embedding), a series of methods have been suggested for incorporating positional embeddings on graph methods \citep{kim2022pure,dwivedi2021graph}. More specifically, \citep{dwivedi2021graph} proposes the MLPGNN-LSPE architecture, which can simultaneously learn both structural and positional embeddings for nodes, whereas \citep{kim2022pure} proposes TokenGT, which treats all nodes and edges as independent tokens, augments them with positional embeddings --  eigenvectors of the normalized Laplacian of the graph ignoring edge directions -- and then feeds them to a Transformer model. 
However, creating positional embeddings that require the factorization of the Laplacian matrix is impractical for large-scale graphs. For this reason, various methods have been proposed for embedding nodes in a graph in a scalable manner, such as TransE \citep{bordes2013translating}, DistMult \citep{yang2014embedding}\footnote{For more such methods see \citep{lerer2019pytorch} and the references therein.}. The recent development of Pytorch-Biggraph \citep{lerer2019pytorch} allows training KGEs large heterogeneous graphs \citep{el2022twhin}.

\end{document}

%% file: glinkx_fast_2_hyperparameters.tex
\begin{table}[h]
\centering
\footnotesize
\begin{tabular}{lrrrrr}
\toprule
 name       &   init\_layers\_A &   init\_layers\_X &   init\_layers\_agg &   inner\_dropout &    lr \\
\midrule
 arxiv-year &                      1 &                      2 &                        1 &                    0.5 & 0.001 \\
 PubMed     &                      1 &                      2 &                        1 &                    0.5 & 0.001 \\
 squirrel   &                      2 &                      1 &                        1 &                    0.5 & 0.001 \\
 yelp-chi   &                      2 &                      2 &                        1 &                    0.5 & 0.01  \\
\bottomrule
\end{tabular}
\caption{GLINKX w/ Adjacency Hyperparameters}
\end{table}

%% file: glinkx_fast_2_biggraph_hyperparameters.tex
\begin{table}[h]
\scriptsize
\begin{tabular}{lrrrrrr}
\toprule
 name          &   biggraph\_vector\_length &   init\_layers\_A &   init\_layers\_X &   init\_layers\_agg &   inner\_dropout &    lr  \\
\midrule
 arxiv-year    &                      400 &                      2 &                      1 &                        1 &                    0.5 & 0.01   \\
 ogbn-arxiv    &                      400 &                      2 &                      2 &                        2 &                    0.5 & 0.001                         \\
 PubMed        &                      400 &                      2 &                      2 &                        2 &                    0.5 & 0.01                          \\
 squirrel      &                      400 &                      2 &                      1 &                        2 &                    0.5 & 0.001                        \\
 yelp-chi      &    400                       &                      2 &                      2 &                        2 &    0.5                    & 0.01                          \\
\bottomrule
\end{tabular}
\caption{GLINKX w/ Biggraph Hyperparameters}
\end{table}

%% file: gat_hyperparameters.tex
\begin{table}[h]
\centering
\footnotesize
\begin{tabular}{lrrrr}
\toprule
 name          &   gat\_heads &   gat\_hidden\_channels &   gat\_num\_layers &    lr \\
\midrule
 arxiv-year    &           4 &                     8 &                1 & 0.1   \\
 ogbn-arxiv    &           4 &                     4 &                1 & 0.1   \\
 PubMed        &           4 &                    16 &                1 & 0.1   \\
 squirrel      &           4 &                     4 &                1 & 0.1   \\
\bottomrule
\end{tabular}
\caption{GAT w/ 1 layer Hyperparameters}
\end{table}

%% file: gcn_hyperparameters.tex
\begin{table}[h]
\centering
\footnotesize
\begin{tabular}{lrrr}
\toprule
 name          &   gcn\_hidden\_channels &   gcn\_num\_layers &    lr \\
\midrule
 arxiv-year    &                    64 &                1 & 0.01  \\
 ogbn-arxiv    &                    64 &                1 & 0.01  \\
 PubMed        &                    64 &                1 & 0.01  \\
 squirrel      &                    64 &                1 & 0.01  \\
 yelp-chi      &                    64 &                1 & 0.01  \\
\bottomrule
\end{tabular}
\caption{GCN w/ 1 layer Hyperparameter}
\end{table}